\documentclass{article}

\usepackage[preprint]{neurips_2026}

\usepackage[utf8]{inputenc} 
\usepackage[T1]{fontenc}    
\usepackage{hyperref}       
\usepackage{url}            
\usepackage{booktabs}       
\usepackage{amsfonts}       
\usepackage{nicefrac}       
\usepackage{microtype}      
\usepackage{xcolor}         

\usepackage{latexsym}

\usepackage{inconsolata}

\usepackage{graphicx}

\usepackage{amsmath,amsfonts,bm}

\def\eqref#1{equation~\ref{#1}}

\def\1{\bm{1}}

\DeclareMathAlphabet{\mathsfit}{\encodingdefault}{\sfdefault}{m}{sl}
\SetMathAlphabet{\mathsfit}{bold}{\encodingdefault}{\sfdefault}{bx}{n}

\usepackage[english]{babel}

\usepackage{svg}
\usepackage{graphicx}

\usepackage{relsize}
\usepackage{nccmath}

\usepackage{pgf}
\usepackage{xcolor}

\usepackage{color, colortbl}

\usepackage{multirow}
\usepackage{dsfont}
\usepackage{wrapfig}
\usepackage{caption}
\usepackage{adjustbox}
\usepackage[format=hang]{subcaption}
\usepackage{soul}

\usepackage{enumitem}
\setlist{nosep}

\def\thebibliography#1{\vskip\parskip%
\vskip\baselineskip%
\def\baselinestretch{1}%
\vskip-\parskip%
\vskip-\baselineskip%
\section*{References}\list
 {}{\setlength{\labelwidth}{0pt}\setlength{\leftmargin}{\parindent}
 \setlength{\itemindent}{-\parindent}}
 \def\newblock{\hskip .11em plus .33em minus -.07em}
 \sloppy\clubpenalty4000\widowpenalty4000
 \sfcode`\.=1000\relax}

\definecolor{darkblue}{rgb}{0, 0, 0.5}
\hypersetup{colorlinks=true, citecolor=darkblue, linkcolor=darkblue, urlcolor=darkblue}

\definecolor{LightRed}{HTML}{f4cfc9}

\newcommand{\rmv}[1]{}

\newcommand{\modelname}{Routing-Free MoE}

\makeatletter
\newcommand{\gradcell}[3]{
  \pgfmathparse{min(100,max(0,(#3-#1)/(#2-#1)*150))}%
  \edef\x{\noexpand\cellcolor{LightRed!\pgfmathresult!white}}\x #3}
\makeatother

\makeatletter
    \def\tagform@#1{\maketag@@@{\normalsize(#1)\@@italiccorr}}
\makeatother

\title{
Routing-Free Mixture-of-Experts}

\author{
Yilun Liu${}^{*,\dagger,1,3}$, Jinru Han${}^{*,2}$, Sikuan Yan${}^{1}$, Volker Tresp${}^{1,3}$, Yunpu Ma${}^{\dagger,1,3}$\\
${}^{1}$ Ludwig Maximilian University of Munich \quad ${}^{2}$ University of California, Los Angeles \\
${}^{3}$ Munich Center for Machine Learning \quad
${}^{*}$ \textit{Equal contribution.}\\
  \small{$^\dagger$ \texttt{yilun.liu@tum.de} \quad
  \texttt{cognitive.yunpu@gmail.com} }
}
\begin{document}
\maketitle

\begin{abstract}
Standard Mixture-of-Experts (MoE) models rely on centralized routing mechanisms that introduce rigid inductive biases.
We propose Routing-Free MoE which eliminates any hard-coded centralized designs including external routers, Softmax, TopK and load balancing, instead encapsulating all activation functionalities within individual experts and are directly optimized through continuous gradient flow, enabling each expert to determine its activation entirely on its own.
We introduce a unified adaptive load-balancing framework to simultaneously optimize both expert-balancing and token-balancing objectives through a configurable interpolation, allowing flexible and customizable resource allocation. 
Extensive experiments show that Routing-Free MoE can consistently outperform baselines with better scalability and robustness.
We analyze its behavior in detail and offer insights that may facilitate future MoE design and optimization.
Code is available at \url{https://github.com/liuyilun2000/RoutingFreeMoE/tree/release}.
\rmv{\url{https://anonymous.4open.science/r/RoutingFreeMoE-8065/}.
}
\end{abstract}



\begin{figure*}[h]
\vspace{0.8em}
    \centering
    \includegraphics[width=.99\linewidth]{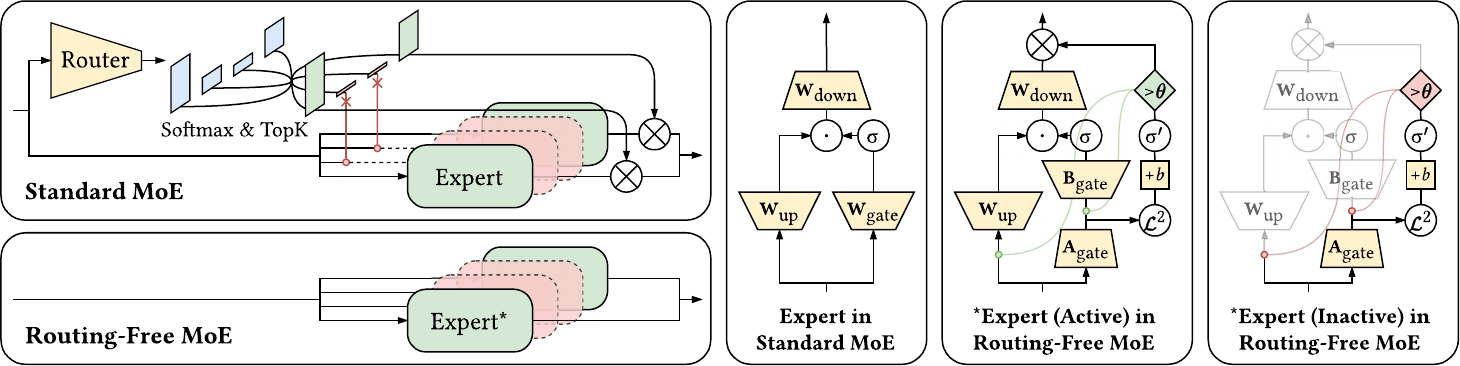}
        \caption{Standard MoE relies on routing to orchestrate expert activations. Routing-Free MoE let each expert purely-independently determine its own activation. Green indicates activated components; red for inactive components; yellow for trainable components.}
    \label{fig:arch}
\end{figure*}
\section{Introduction}
\rmv{
our story here as conventional MoE is regarding modeling optimal expert distribution as an **top-down** **classification** to **distribute** experts/tokens, and our is **bottom-up** that allows **emergence of activation patterns**.
}
The scalability of transformer-based Large Language Models (LLMs) \citep{vaswani2017attention} is constrained by the substantial computational resources required \citep{kaplan2020scaling}.
To efficiently expand model capacity without proportional computation cost growth, Mixture-of-Experts (MoE) designs focus on activating subsets of parameters for each input \citep{shazeer2017outrageously, fedus2022switchtransformer}. 
This approach presents a fundamental challenge: how to optimally distribute inputs to experts while satisfying sparsity and balancing considerations.

Existing MoE designs are hindered by structural limitations across multiple dimensions.
Standard MoE relies on small, external routers that lack sufficient capacity for storing expert capabilities to determine expert preference for each input, forcing them to learn their prediction via indirect trial-and-error optimization, which inevitably leads to suboptimal routing and unstable early training dynamics \citep{lv2025aoe}.
For computational efficiency, conventional MoE enforces rigid, global constraints on expert activation that ignore input-specific dynamics.
The fixed TopK selection imposes uniform sparsity, regardless of varying input complexity and expertise of experts, overlooking potential gains from fewer or more activations \citep{zhou2022mixture}.
The Softmax operation forces a competitive probability distribution that sacrifices the absolute magnitude information of expert activations \citep{wang2024remoe}.
\rmv{here about activation PATTERN}
For load-balancing, existing strategies employ different balancing targets that force predetermined activation patterns, which are often mutually exclusive yet exhibit varying performance under different configurations \citep{fedus2022switchtransformer, zhou2022mixture, muennighoff2024olmoe}.
Rigidly adhering to either may constrain the model's ability to adaptively discover potentially better resource allocation patterns \citep{do2025usmoe}.

To bridge these gaps, we introduce Routing-Free MoE, a bottom-up architecture without any centralized routers, Softmax, TopK, or rigid load-balancing designs, where each expert individually and directly determines its own activation. 
Our design enables each expert to activate itself purely when its internal confidence score surpasses a configurable threshold.
To satisfy efficiency requirements, we design a dynamic, configurable framework to adaptively achieve the sparsity and load-balancing objectives during training, allowing the optimal activation pattern to emerge spontaneously.
We utilize an auxiliary loss function that seamlessly integrates both token and expert balancing, providing flexibility to adaptively exploit both depending on training dynamics and workload requirements.

We validate Routing-Free MoE across three scales up to 0.8B with extensive experiments in comparison against standard MoE and strong baselines.
Across all settings, Routing-Free MoE consistently achieves better language modeling quality and downstream performance averaged across 9 evaluation benchmarks.
It also demonstrates notably improved scalability and robustness.
We analyze its training behavior in detail and further document intriguing phenomena for density and load-balancing, offering insights that may guide future improvements in MoE design and optimization.

\begin{figure}[t]
    \centering
    \includegraphics[width=.4\linewidth]{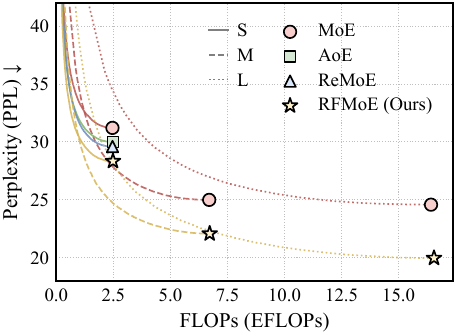}
        \caption{Routing-Free MoE consistently outperforms standard MoE, AoE \citep{lv2025aoe}, and ReMoE \citep{wang2024remoe} in language modeling. All models are trained on OpenWebText \citep{Gokaslan2019OpenWeb} under identical environment conditions and best-performing configurations, as described in Section \ref{sec:exp-setup}. FLOPs are estimated for one epoch.}
    \label{fig:PPL-result}
\end{figure}

In summary, our key contributions are:

\begin{itemize}[leftmargin=*]
  \item A Routing-Free MoE architecture that eliminates routers, Softmax, TopK, and hard-coded load balancing mechanisms.
  \item A unified, adaptive load-balancing framework that jointly optimizes token-balancing and expert-balancing, allowing customizable application.
  \item Experiments and analyses demonstrating the improvements of Routing-Free MoE over baselines.
\end{itemize}

\section{Preliminaries}
Consider a standard MoE LLM\citep{fedus2022switchtransformer, jiang2024mixtral} comprising $L$ transformer blocks.\rmv{, each with a block of standard self-attention followed by a MoE feed-forward neural network (FFN) block. }
For layer $\ell \in \{1,\cdots,L\}$ and token sequence of length $T$, the forward pass\footnote{LayerNorms and dropout are omitted for clarity.} can be formulated as: 

\begin{align}
\mathbf{x}^\ell_{1:T} &= \mathrm{SelfAttn}(\mathbf{h}^{\ell-1}_{1:T}) + \mathbf{h}^{\ell-1}_{1:T}, \\
\mathbf{h}^\ell_t &= \mathrm{MoE}(\mathbf{x}^\ell_t) + \mathbf{x}^\ell_t,
\end{align}
where $\mathbf{x}^\ell_{1:T}$ denotes the attention module output with residual connection added. The MoE block performs a token-wise mapping, yielding output $\mathbf{h}^\ell_t$ at token $t\in \{1,\cdots,T\}$ with residual added. 

The mainstream form of MoE LLMs incorporates multiple structurally identical FFN experts $E_i(\cdot), i\in\{1,\cdots,N\}$ within the MoE layer.
This is implemented via a token-wise routing mechanism among all $N$ experts at layer $\ell$, with a gating network $G(\cdot)$ assigning each token to a designated number $K$ of top-activated experts:
\begin{align}
    \mathbf{h} &= \sum\nolimits^N_{i=1}\left(G\left(\mathbf{x}\right)_i E_i\left(\mathbf{x}\right)\right) + \mathbf{x}, \\
    G\left(\mathbf{x}\right) &= \mathrm{Softmax}\left(\mathrm{TopK}\left(\mathbf{x}\mathbf{G}, K\right)\right), \label{eq:moe}
\end{align}
where $G(\cdot): \mathbb{R}^D \mapsto \mathbb{R}^N $ denotes the routing mechanism, whose output serves as the weights of the weighted sum for outputs among all $N$ experts, and only $K$ of $N$ experts receive nonzero values.
The router learns its weight matrix $\mathbf{G} \in \mathbb{R}^{D \times N}$ that can be interpreted as a set of $N$ individual $D$-dimensional expert vectors $\mathbf{g}_i$, each responding to a characteristic hidden state $\mathbf{h}_i$ that should activate the corresponding expert $E_i$, with $\mathbf{s}_i=\mathbf{xg}_i\in \mathbb{R}^N$ denoting the token-to-expert affinity scores \citep{zhou2022mixture, liu2025parameter}. 
$\mathrm{TopK}(\mathbf{s},K)$ retains the top-$K$ scores and masks the rest to $-\infty$.
\rmv{
$\mathrm{TopK}(\mathbf{s},K)$ selects of $K\le N$ highest scores in $\mathbf{s}$:
\begin{equation}
    \mathrm{TopK}(\mathbf{s})_i = 
    \begin{cases} 
        \mathbf{s}_i, &i \in \arg\mathrm{topK}(\mathbf{s}), \\
        -\infty, & \mathrm{otherwise}.
    \end{cases} 
\end{equation}
}

Each individual expert $E_i$ is implemented as a Feed-Forward Network (FFN).
Modern FFN typically takes the form of a Gated Linear Unit (GLU, \citet{dauphin2017language, shazeer2017outrageously}): 

\begin{equation}
\label{eq:GLU}
\mathrm{FFN}(\mathbf{x})=[\sigma(\mathbf{x}\mathbf{W}_\mathrm{up})\odot(\mathbf{x}\mathbf{W}_\mathrm{gate})]\mathbf{W}_\mathrm{down},    
\end{equation}
where $\sigma$ is the activation function and $\odot$ denotes element-wise multiplication. 
In MoE LLMs, this design enables a second level of input-dependent information filtering, leading to potentially more effective representations.

The routing in standard MoE relies on several strong inductive biases.
With merely $N$ expert vectors $\mathbf{g}_i\in\mathbb{R}^D$, the router's capacity is orders of magnitude smaller than the experts themselves, yet must compress the knowledge-intensive activation preferences of all $N$ experts into single dot-product scores without any direct signal about expert capabilities, improving only through indirect trial-and-error \citep{lv2025aoe}. 
TopK hard-codes a fixed sparsity ratio $K/N$ regardless of input complexity, preventing input-adaptive activation patterns \citep{zhou2022mixture}. 
Softmax discards absolute activation magnitudes by forcing a competitive probability distribution, suppressing the residual contribution of highly suitable experts when others also happen to score higher \citep{wang2024remoe}.

\section{Methodology}
\subsection{Architecture}

To alleviate the router bottleneck, \citet{lv2025aoe} introduce Autonomy-of-Experts (AoE), with
\begin{equation}
\mathrm{FFN}(\mathbf{x})=[\sigma(\mathbf{x}\mathbf{A}_{\mathrm{gate}}\mathbf{B}_{\mathrm{gate}})\odot(\mathbf{x}\mathbf{W}_{\mathrm{up}})]\mathbf{W}_{\mathrm{down}},
\label{eq:aoe-ffn}\end{equation}
where $\mathbf{A}_\mathrm{gate}\in \mathbb{R}^{D\times r}$ projects the input hidden state $\mathbf{x}$ from hidden dimension $D$ to a lower-dimensional rank $r \ll D$, and $\mathbf{B}_\mathrm{gate}\in \mathbb{R}^{r\times D_\mathrm{act}}$ projects it back.
This low-rank representation provides an alternative indicator of expert suitability that originates from within the expert itself rather than from an external router.
Each expert can therefore directly produce its own scalar activation score by applying a norm to its own internal representation $\|\mathbf{x}\mathbf{A}_{\mathrm{gate},i}\|_2$.
However, AoE feeds the internal scores $\|\mathbf{xA}_{\mathrm{gate}}\|_2$ back to the standard centralized TopK and Softmax routing pipeline:
\begin{align}
    G(\mathbf{x})=\mathrm{Softmax}(\mathrm{TopK}(\|\mathbf{xA}_\mathrm{gate}\|_2, K)),
\end{align}
thereby retaining the structural constraints and inductive biases of conventional routing.

\rmv{
However, AoE's architectural design stops halfway at replacing $\mathbf{xG}$ with $\|\mathbf{xA}_{\mathrm{gate}}\|_2$, which in their design is subsequently fed back into the standard centralized routing pipeline for TopK and Softmax calculations.
The decision of which experts to activate still happens externally with all experts' $\|\mathbf{xA}_{\mathrm{gate},i}\|_2$ collected and compared with each other, and the decisions are sent back to the activated experts to continue their calculation of remaining FFN modules.
Such design still preserves the structural constraints that impose artificial inductive biases on expert activation.
}

Meanwhile, addressing the constraints of TopK and Softmax, \citet{wang2024remoe} propose ReMoE, which replaces TopK and Softmax with a single ReLU function applied directly to router's output:
\begin{equation}
G(\mathbf{x}) = \mathrm{ReLU}(\mathbf{x}\mathbf{G}),
\end{equation}
The sparse activation arises naturally from ReLU without any explicit TopK selection or comparative normalization.
The absolute magnitude of router scores is preserved, allowing each expert’s residual contribution to be linearly weighted by router's prediction, rather than a normalized relative preference.
Nevertheless, ReMoE still retains a centralized external router, preserving the information bottleneck and indirect optimization dynamics.

Building on these insights, our Routing-Free MoE seeks to eliminate all constraints of routing mechanisms.
We adopt AoE's FFN design (Equation \ref{eq:aoe-ffn}) using each expert's internal norm $\|\mathbf{xA}_\mathrm{gate}\|_2$ as the initial expert preference score, grounding the activation decision in the expert's own response to the input.
Since $\|\mathbf{xA}_\mathrm{gate}\|_2$ is strictly non-negative unlike router's $\mathbf{xG}$, we introduce a learnable per-expert bias term before activation upon ReMoE's design, yielding
\begin{equation}    
G_i(\mathbf{x})=\mathrm{ReLU}(\|\mathbf{xA}_{\mathrm{gate},i}\|_2-b_i). \label{eq:rf-gate}
\end{equation}
With the per-expert bias term introduced, experts whose weighted norm falls below their own bias threshold contribute zero and are effectively deactivated, allowing each expert to jointly adapt both its $\mathbf{A}_{\mathrm{gate},i}$ matrix and the $b_i$ parameter to effectively adjust its own activation ratio.
We further introduce a global post-activation threshold $\theta$ as a configurable hyperparameter for external control over the overall sparsity level, giving the final binary activation decision of each expert:
\begin{equation}
    f_i(\mathbf{x})=\mathds{1}\left\{G_i(\mathbf{x})- \theta\geq0\right\}. \label{eq:rf-mask}
\end{equation}
The result is a fully decentralized architecture without any external router, TopK or Softmax, making each expert independently determine its own activation from within, allowing the global activation pattern to emerge bottom-up from collective self-adjustment of all experts.
Figure \ref{fig:arch} visualizes the architecture of Routing-Free MoE and its experts.

\begin{figure*}[t]
    \centering
    \begin{subfigure}[b]{0.32\linewidth}
        \centering
        \includegraphics[scale=0.48]{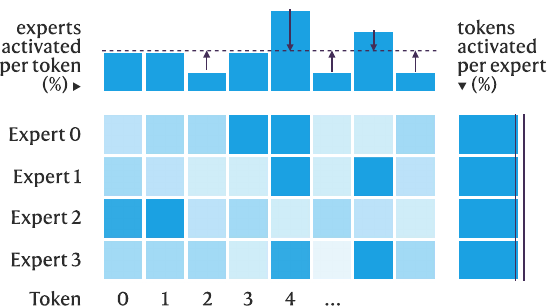}
        \caption{Expert choice (EC) ensures hard expert-balancing and optimizes token-balancing via training.}
        \label{fig:density}
    \end{subfigure}
    \hfill
    \begin{subfigure}[b]{0.32\linewidth}
        \centering
        \includegraphics[scale=0.48]{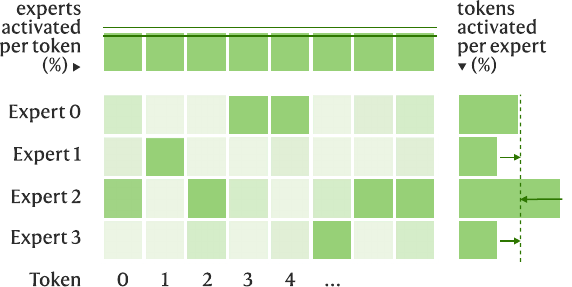}
        \caption{Token choice ensures hard token-balancing and optimizes expert-balancing via training.  }
        \label{fig:lambda_reg}
    \end{subfigure}
    \hfill
    \begin{subfigure}[b]{0.33\linewidth}
        \centering
        \includegraphics[scale=0.48]{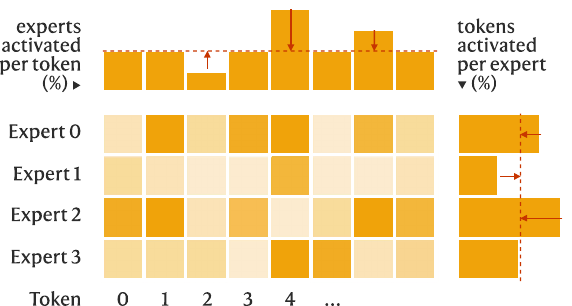}
        \caption{Routing-Free MoE adaptively optimizes both token-balancing and expert-balancing via training.}
        \label{fig:losses}
    \end{subfigure}
    \caption{Load-balancing for tokens and experts.  Routing-Free MoE introduces a unified load-balancing framework that simultaneously optimizes both expert-balancing and token-balancing through a configurable interpolation.}
    \label{fig:load-balancing}
\end{figure*}

\subsection{Training}

Training MoE models requires simultaneously maintaining the activation ratio and balanced expert and token distribution.
Standard practice hard-codes TopK for the activation ratio, and addresses balancing via either \emph{Token-Choice} (TC) \citep{shazeer2017outrageously, fedus2022switchtransformer} that guarantees per-token compute but not expert balance, or \emph{Expert-Choice} (EC) \citep{zhou2022mixture} that hard-codes uniform expert utilization but not per-token compute, as illustrated in Figure~\ref{fig:load-balancing}.
Both enforce one balanced dimension as a hard constraint and optimizes the other via soft auxiliary loss.
\rmv{
When training MoE models for language modeling, a critical challenge is to simultaneously satisfying multiple practical constraints at once, including achieving a specified activation ratio to control the average computational budget, while also maintaining balanced expert and token distribution to ensure efficiency, stability, and robustness.
Standard practice typically hard-codes a TopK to ensure the activation ratio, and addresses load balancing via two different approaches, as illustrated in Figure~\ref{fig:load-balancing}.
The most common paradigm is \emph{Token-Choice} (TC) \citep{shazeer2017outrageously, fedus2022switchtransformer}, 
where each token independently selects its top-$K$ experts via a learned router. 
This guarantees that each token is processed by exactly $K$ experts, but offers no guarantee that tokens 
are evenly spread across experts, which requires auxiliary losses during training to discourage expert collapse.
In \emph{Expert-Choice} (EC) \citep{zhou2022mixture}, each expert instead selects a fixed number of top-scoring tokens from the batch. 
This hard-codes uniform expert utilization, but may leave some tokens unprocessed or processed by fewer experts than others, breaking the guarantee of uniform per-token compute.
Both paradigms thus enforce one dimension of balance as a hard constraint while optimizing the other as a soft objective via auxiliary losses.
}
As Routing-Free MoE has eliminated all centralized routing mechanisms, standard activation ratio and load-balancing strategies that rely on hard-coded TopK \citep{do2025usmoe} no longer apply. We introduce a unified load-balancing framework by extending the auxiliary loss of \citet{fedus2022switchtransformer} to jointly encourage both balanced token distribution across experts and balanced expert activation per token, without requiring any centralized mechanisms.

As the binary activation decision $f_i$ in Equation \ref{eq:rf-mask} is non-differentiable, we directly use the pre-threshold activation weight $G_i(\mathbf{x})$ as a differentiable activation proxy of expert $E_i$.
We define the mean activation density over a set of experts $\mathcal{E}$ and token batch $\mathcal{B}$ and its differentiable proxy as:
\begin{align}
    \rho(\mathcal{E},\mathcal{B}) &= \frac{1}{|\mathcal{E}||\mathcal{B}|}
    \sum_{e_i\in\mathcal{E}}\sum_{\mathbf{x}\in\mathcal{B}} f_i(\mathbf{x}),\\
    \tilde{\rho}(\mathcal{E},\mathcal{B}) &= \frac{1}{|\mathcal{E}||\mathcal{B}|}
    \sum_{e_i\in\mathcal{E}}\sum_{\mathbf{x}\in\mathcal{B}} G_i(\mathbf{x}).
\end{align}
Both quantities equal the target activation density $\rho_\infty$ under perfectly uniform load.

We decompose the load-balancing objective into two complementary terms\footnote{We avoid using the terms expert choice and token choice in Routing-Free MoE because the notion of ``choice'' implies a centralized comparison and selection process, which  contradicts our principle that all decisions should emerge locally and independently at individual experts and tokens.}. 
The \emph{expert-balancing} loss $\mathcal{L}_\mathrm{EB}$ encourages uniform distribution  of tokens across experts by penalizing experts that consistently receive more or fewer tokens than average:
\begin{equation}
    \mathcal{L}_\mathrm{EB} = \frac{1}{|\mathcal{E}|}\sum_{e_i\in\mathcal{E}}
    \left(\frac{1}{|\mathcal{B}|}\sum_{\mathbf{x}\in\mathcal{B}} f_i(\mathbf{x})\right)
    \left(\frac{1}{|\mathcal{B}|}\sum_{\mathbf{x}\in\mathcal{B}} G_i(\mathbf{x})\right).
\end{equation}
The \emph{token-balancing} loss $\mathcal{L}_\mathrm{TB}$ encourages uniform distribution 
of activated experts per token by penalizing tokens that consistently activate more or fewer experts than average:
\begin{equation}
    \mathcal{L}_\mathrm{TB} = \frac{1}{|\mathcal{B}|}\sum_{\mathbf{x}\in\mathcal{B}}
    \left(\frac{1}{|\mathcal{E}|}\sum_{e_i\in\mathcal{E}} f_i(\mathbf{x})\right)
    \left(\frac{1}{|\mathcal{E}|}\sum_{e_j\in\mathcal{E}} G_j(\mathbf{x})\right).
\end{equation}
Each loss is a dot product between a binary non-differentiable term and a differentiable proxy, and is minimized when both factors equal $\rho_\infty$ uniformly\footnote{See Appendix \ref{app:lb_theory}.}. 
The two objectives are combined via a configurable interpolation parameter $\mu \in [0, 1]$:
\begin{equation}
    \mathcal{L}_\mathrm{LB} = \mu\,\mathcal{L}_\mathrm{EB} + (1-\mu)\,\mathcal{L}_\mathrm{TB}.
\end{equation}
Setting $\mu = 1$ recovers a pure expert-balancing, and $\mu = 0$ 
recovers pure token-balancing.
This provides a single unified framework that interpolates both routing paradigms, allowing load-balancing to be tailored to specific deployment needs.

Rather than fixing the auxiliary loss coefficient $\lambda$ as a static hyperparameter, we follow \citet{wang2024remoe} and
adopt an adaptive schedule that drives the empirical activation ratio toward the target $\rho_\infty$ throughout training. The total training 
objective is
\begin{equation}
    \mathcal{L} = \mathcal{L}_\mathrm{LM} + \lambda_t\,\mathcal{L}_\mathrm{LB},
\end{equation}
with $\lambda_t$ updated at each training step $t$ as:
\begin{equation}
    \lambda_{t+1} = \lambda_t \cdot \left(1 + \eta\right)^{\mathrm{sign}
    \left(\rho_t(\mathcal{E},\mathcal{B}) - \rho_\infty\right)}.
\end{equation}
When the current density $\rho_t$ exceeds the target $\rho_\infty$, $\lambda_t$ 
increases to exert stronger load-balancing pressure; when below, it decreases to allow more expert activations. 
The step size $\eta$ controls the responsiveness of this feedback loop. This formulation avoids the need to manually tune $\lambda$ while ensuring the model converges to the desired computational budget, and is compatible with both training from scratch and adaptation of pretrained MoE models into Routing-Free MoE.

To encourage all experts to participate during the early warm-up phase of training, each expert’s bias is initialized as $1\mathrm{e}^{-6}$, allowing all experts to become activated, jointly exploring the representation space and establishing initial specialization before sparsity is enforced.
As training progresses and
$\lambda$  increases, the sparsity regularization gradually strengthens, driving the activation density toward target $\rho_\infty$. 
By this stage, experts have already developed distinct roles, avoiding expert collapse and ensuring a stable and effective phase transition.

\rmv{
where $f_i$ is the fraction of tokens dispatched to expert $i$
\begin{equation}
    f_i(\mathcal{B})=\frac{1}{T}\sum_{\mathbf{x}\in\mathcal{B}}\mathds{1}\{i\in \arg\mathrm{TopK}(p(\mathbf{x}))\}
\end{equation}
$P_i$ is the fraction of the router probability allocated for expert $i$
\begin{equation}
    p_i(\mathcal{B})=\frac{1}{T}\sum_{\mathbf{x}\in\mathcal{B}}p_i(\mathbf{x})
\end{equation}

\begin{equation}
    f(x)=\frac{1}{N}\sum_{i=1}^{N}\mathds{1}[i\in \mathrm{TopK}(p(\mathbf{x}))]
\end{equation}

\begin{equation}
    p(\mathbf{x})=\frac{1}{N}\sum_{i=1}^{N}p_i(\mathbf{x})
\end{equation}
}

\begin{table*}[t]
\caption{Evaluation results for standard MoE baseline and \modelname{} across scales and benchmarks. Each model is trained on OpenWebText \citep{Gokaslan2019OpenWeb} from scratch for one epoch. Details in Appendix \ref{appdix:result}. \rmv{Dataset names are partially abbreviated, including PIQA \citep{bisk2020piqa}, HellaSwag \citep{zellers2019hellaswag}, WinoGrande \citep{sakaguchi2021winogrande}, ARC-Easy and ARC-Challenge \citep{clark2018arc}, OpenBookQA \citep{mihaylov2018obqa}, Quora Question Pairs \citep{wang2018glue}, Question NLI \citep{wang2018glue}, and SST-2 \citep{wang2018glue}. 
}}
\label{tab:main_result}
\begin{adjustbox}{width=\textwidth}
    \small\addtolength{\tabcolsep}{-2pt} 
    \begin{tabular}{c|r|cc|c|ccccccccc|c}
    \toprule
        & \textbf{Arch.}
            & \textbf{Size} & \textbf{FLOPs}
            & \textbf{PPL$\downarrow$}
            & \textbf{PIQA} & \textbf{HellaS} & \textbf{WinoG}
            & \textbf{ARCe} & \textbf{ARCc}
            & \textbf{OBQA} & \textbf{QQP}
            & \textbf{QNLI}  & \textbf{SST-2}
            & \textbf{Avg.} \\
    \midrule     
        \multirow{4}{*}{S}
        & MoE   
            & 92.44M & 90.93M & 31.22 &
            57.56 & 27.19 & 51.22 & 33.42 & 21.33 & 24.60 & 36.82 & 49.46 & 49.08 & 38.96 \\
        & AoE 
            & 93.85M & 88.57M & 30.00 & 
            56.09 & 27.01 & 50.20 & 33.84 & 21.93 & 25.00 & 36.82 & 49.46 & 49.08 & 38.82 \\
        & ReMoE 
            & 92.44M & 90.93M & 29.60 &
            56.58 & 26.69 & 51.62 & 33.38 & 22.27 & 26.00 & 36.83 & 49.48 & 49.08 & 39.10 \\
        & RFMoE 
            & 95.32M & 91.08M & 27.42 &
            58.49 & 27.09 & 50.59 & 35.77 & 21.59 & 24.40 & 36.98  & 49.44 & 53.56 & \textbf{39.77} \\
    \midrule
        \multirow{2}{*}{M}
        & MoE
            & 289.9M & 248.0M & 25.00
            & 58.32 & 28.26 & 52.33
            & 36.20 & 21.42
            & 24.60 & 36.85
            & 49.46 & 49.31
            & 39.64 \\
        & RFMoE
            & 307.3M & 249.2M & 22.08
            & 58.92 & 27.85 & 49.72
            & 35.27 & 21.50
            & 26.40 & 37.06
            & 49.51 & 57.34
            & \textbf{40.40} \\
    \midrule
        \multirow{2}{*}{L}
        & MoE
            & 808.4M & 608.4M & 24.58
            & 59.19 & 29.37 & 50.83
            & 36.41 & 23.29
            & 25.00 & 37.57
            & 49.28 & 49.08
            & 40.00 \\
        & RFMoE
            & 870.6M & 613.2M & 19.97
            & 58.87 & 28.27 & 50.51
            & 37.46 & 21.67
            & 26.60 & 39.93
            & 49.86 & 53.67
            & \textbf{40.76} \\
    \bottomrule
    \end{tabular}
\end{adjustbox}
\end{table*}

\section{Experiments}

We conduct comprehensive experiments to validate the performance of Routing-Free MoE.

\subsection{Experimental Setup}
\label{sec:exp-setup}

We implement Routing-Free MoE upon the HuggingFace implementation\footnote{\url{https://github.com/huggingface/transformers/tree/v4.57.6/src/transformers/models/mixtral}} for the Mixtral architecture \citep{jiang2024mixtral}, which includes attention designed as Grouped Query Attention (GQA, \citet{ainslie2023gqa}) with Rotary Position Embeddings (RoPE, \citet{su2024roformer}), FFN using SwiGLU activation \citep{shazeer2020glu}, and Root Mean Square Layer Normalization (RMSNorm, \citet{zhang2019root}) applied to the residual stream prior to each attention and MoE FFN \citep{xiong2020layer}. 

We conduct experiments across three scales (S/M/L) up to 0.8B, and compare against a standard MoE baseline with identical structure except for the routing mechanism. 
AoE \citep{lv2025aoe} and ReMoE \citep{wang2024remoe} are also examined as both additional baselines and ablated variants for comparison.
Detailed hyperparameters and training configurations are in Table \ref{tab:all-result} in Appendix.

Following standard practices, all models are trained on OpenWebText \citep{Gokaslan2019OpenWeb} for one epoch. 
Zero-shot evaluation is conducted across 9 benchmarks \citep{gao2021framework}.
We include PIQA \citep{bisk2020piqa}, HellaSwag \citep{zellers2019hellaswag}, WinoGrande \citep{sakaguchi2021winogrande}, ARC-Easy and ARC-Challenge \citep{clark2018arc}, OpenBookQA \citep{mihaylov2018obqa}, and QQP, QNLI, and SST-2 from GLUE \citep{wang2018glue}. Accuracy is reported on WinoGrande, QQP, QNLI, and SST-2, and normalized accuracy on others.

\subsection{Main Results}
Table~\ref{tab:main_result} presents language modeling perplexity and downstream benchmark evaluation results across all three scales under an iso-compute comparison, with FLOPs matched within $\sim$1\% between MoE and  Routing-Free MoE at each scale. 
Figure~\ref{fig:PPL-result} shows the detailed validation perplexity evolution with total FLOPs throughout training.
Routing-Free MoE achieves consistently better validation perplexity than standard MoE across all scales. 
On downstream benchmarks, Routing-Free MoE also consistently improves the average performance, with a detailed statistical analysis (Appendix \ref{app:stat_significance}) supporting that Routing-Free MoE can produce a statistically significant improvement over the standard MoE baseline.
Notably, the gain of Routing-Free MoE does not diminish with scale, suggesting that the benefits of our approach can continue to provide gains as models scale up.

\begin{figure*}[t]
    \centering
    \begin{subfigure}[b]{0.32\linewidth}
        \centering
        \includegraphics[height=3.2cm]{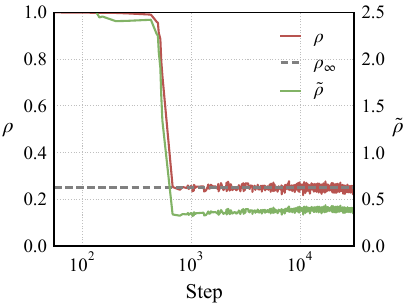}
        \caption{Density $\rho$, density target $\rho_\infty$, and density proxy $\tilde{\rho}$}
        \label{fig:density}
    \end{subfigure}
    \hfill
    \begin{subfigure}[b]{0.32\linewidth}
        \centering
        \includegraphics[height=3.2cm]{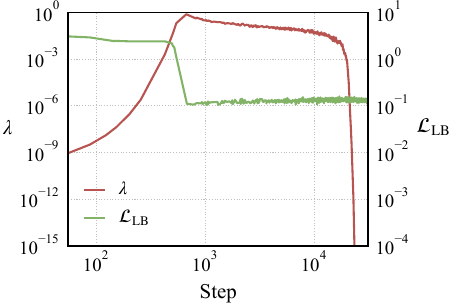}
        \caption{Adaptive coefficient $\lambda$ and load-balancing loss $\mathcal{L}_\mathrm{LB}$ }
        \label{fig:lambda_reg}
    \end{subfigure}
    \hfill
    \begin{subfigure}[b]{0.32\linewidth}
        \centering
        \includegraphics[height=3.2cm]{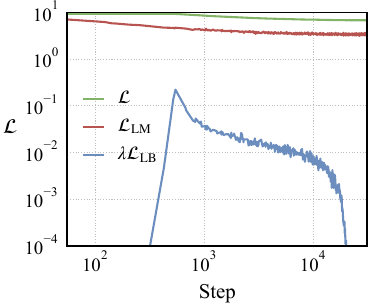}
        \caption{Training $\mathcal{L}$, language modeling $\mathcal{L}_\mathrm{LM}$, and regularization $\lambda\mathcal{L}_\mathrm{LB}$}
        \label{fig:losses}
    \end{subfigure}
    \caption{Training dynamics of Routing-Free MoE at scale S, with $r=16$, $\lambda_0=1\mathrm{e}^{-10}$, $\eta=0.02$, and $\alpha=1\mathrm{e}^{-3}$.}
    \label{fig:loss_metrics}
\end{figure*}

\begin{table}[t]
    \caption{Ablation performed at scale S. Detailed configurations for each run are provided in Table \ref{appdix:result}.}
    \label{tab:ablation}
\vspace{0.8em}
    \centering
    \small
    \addtolength{\tabcolsep}{-2pt} 
    \begin{tabular}{l|cc|c}
\toprule
        \textbf{Config.} & \textbf{Size} & \textbf{FLOPs} & \textbf{PPL$\downarrow$}\\
\midrule
        Standard MoE & 92.44M & 90.93M & 31.22\\
        w/o router (AoE, $r$=16) & 93.85M & 88.57M & 30.00 \\
        w/o router (AoE, $r$=32) & 95.32M & 91.08M & 30.31\\
        w/o TopK\&Softmax (ReMoE) & 92.44M & 90.93M & 29.60\\
        Routing-Free MoE ($r$=16)& 93.85M & 88.57M & 28.73\\
        Routing-Free MoE ($r$=32)& 95.32M & 91.08M & 28.33\\
\bottomrule
    \end{tabular}
\end{table}

\subsection{Architectural Comparison}

We experiment at scale S, decomposing the contribution of each architectural change over the standard MoE via incremental ablation, with main results shown in Table~\ref{tab:ablation}.
\rmv{Replacing the external router with AoE's internal norm-based scoring alone yields a perplexity reduction up to 1.22-point.
Independently, ReMoE's replacing TopK and Softmax with ReLU yields a 1.62-point reduction.}
Routing-Free MoE outperforms either by a substantial margin under matching size and FLOPs.
A visualization of perplexity curves during training for AoE and ReMoE can also be found in Figure \ref{fig:PPL-result}.

\begin{table}[t]
\caption{Effect of low-rank projection dimension at scale S, with learning rate $\alpha=1\mathrm{e}^{-3}$. }
\label{tab:ablation_rank}
\vspace{0.8em}
\centering
\small
\begin{tabular}{r|cc|c}
\toprule
\textbf{$\bm{r}$} & \textbf{Size} & \textbf{FLOPs} & \textbf{PPL$\downarrow$} \\
\midrule
8 & 93.11M & 87.32M & 29.16 \\
16 & 93.85M & 88.57M & 28.74 \\
32 & 95.32M & 91.08M & 28.34 \\
64 & 98.27M & 96.09M & 28.24\\
\bottomrule
\end{tabular}
\end{table}

Table~\ref{tab:ablation_rank} reports the effect of $r$. Increasing $r$ consistently improves perplexity, yet with diminishing returns. Therefore, we set 32 as the default $r$ and scale it proportionally with the hidden dimension $D$ across experiments on other scales for the optimal efficiency--quality trade-off.

A detailed analysis about Routing-Free MoE at deployment is provided in Appendix \ref{app:deployment}, including efficiency improvements by expert parallelism and the effects of adapting threshold $\theta$ at inference.

\subsection{Training Dynamics}

Figure~\ref{fig:loss_metrics} visualizes the training dynamics of \modelname{}.
The empirical activation density $\rho$ begins near 1 at
initialization, drops sharply to the target $\rho_\infty$ as $\lambda$ grows, and remains close to the target thereafter.
The differentiable proxy $\tilde{\rho}$ also settles slightly above $\rho_\infty$ at convergence.
$\lambda$ rises steeply during the initial warm-up phase, plateaus as $\rho$ converges to $\rho_\infty$, and collapses sharply near the end as density falls marginally below the target.
The regularization term $\lambda\mathcal{L}_\mathrm{LB}$ exhibits a transient spike caused by the exponential growth of $\lambda$, with its magnitude rising above $10^{-1}$, making it a non-negligible contributor to the total loss $\mathcal{L}$ alongside $\mathcal{L}_\mathrm{LM}$, and steering gradient descent toward directions that also reduce $\mathcal{L}_\mathrm{LB}$.
As training progresses, $\lambda\mathcal{L}_\mathrm{LB}$ decays to a negligible level, demonstrating how load‑balancing pressure is applied precisely when needed without persistently distorting the language‑modeling optimization objectives.

\begin{figure}[t]
    \centering
    \begin{subfigure}[b]{0.49\linewidth}
        \centering
        \includegraphics[width=\linewidth]{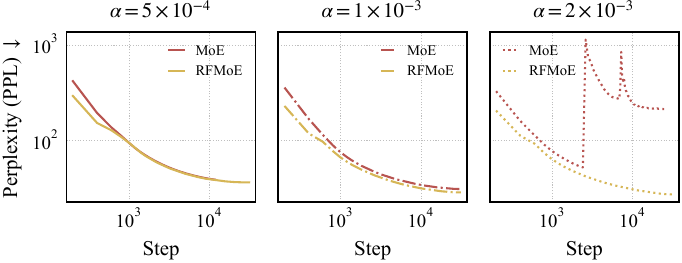}
        \caption{Training curves with different $\alpha$ at scale S}
        \label{fig:density}
    \end{subfigure}
    \hfill
    \begin{subfigure}[b]{0.49\linewidth}
        \centering
        \includegraphics[width=\linewidth]{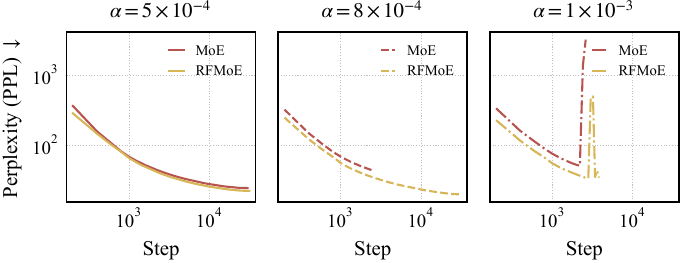}
        \caption{Training curves with different $\alpha$ at scale L}
        \label{fig:lambda_reg}
    \end{subfigure}
    \caption{Training dynamics of Routing-Free MoE by $\alpha$, with $r=16$, $\lambda_0=1\mathrm{e}^{-10}$, and $\eta=0.02$.}
    \label{fig:lr}
\end{figure}

Figure \ref{fig:lr} further examines training stability by sweeping the learning rate $\alpha$. Under conservative $\alpha$, both models converge smoothly with \modelname{} consistently achieving lower perplexity. As $\alpha$ increases, the standard MoE baseline begins to collapse much earlier than \modelname{}. This is evident at scale S, where MoE fails at $\alpha=2\mathrm{e}^{-3}$ while \modelname{} remains stable throughout training.
In addition, tuning $\alpha$ for the standard MoE at scale L yields only marginal gains over its scale-M performance (as in Figure \ref{fig:PPL-result}); whereas \modelname{} continues to improve as scale increases,  highlighting its enhanced scalability.
\modelname{} not only outperforms the baseline, but also exhibits better training stability and less sensitivity to learning rates, tolerating a wider range of hyperparameter choices at scale.

\begin{figure*}[t]
    \centering
    \includegraphics[width=\linewidth]{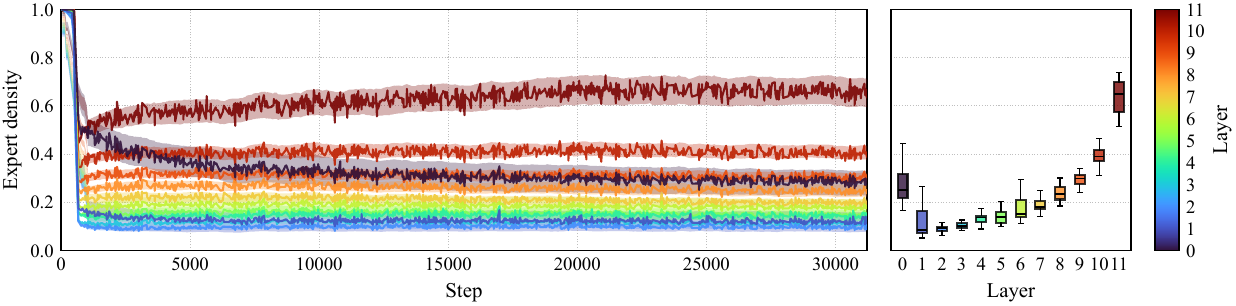}
        \caption{Evolution of expert activation across layers during training on OpenWebText under global density target.  
Left panel shows the per‑layer mean activation as lines, along with IQR as shaded band regions using 1,000‑step moving average smoothing.  
Right panel shows activation distribution at the final training step. Model at scale S. }
    \label{fig:density_layer}
\end{figure*}

\section{Discussion}

Beyond its superior performance compared with the baselines, here we analyze some intriguing properties of Routing-Free MoE.

\subsection{Per-Layer and Global Density}

With Top-K at each layer eliminated, a natural question is whether the target activation density $\rho_\infty$ should still be enforced  within each layer, or only globally across all experts.
Per-layer enforcement computes $\mathcal{L}_{LB}$ separately at each layer, continuing to impose the inductive bias for a uniform sparsity at every depth. 
A global $\rho_\infty$, in contrast, relaxes the constraint and allows individual layers to deviate from $\rho_\infty$ so long as aggregate activation matches the target and improves the overall loss $\mathcal{L}$. 
Our experiment on scale S finds that relaxing this inductive bias leads to a significant improvement with perplexity drops from 39.44 to 28.74.
In Figure \ref{fig:density_layer} we illustrate the emergent expert activation pattern that Routing-Free MoE develops by itself during training when this per-layer bias is removed.
A detailed analysis on this matter is provided in Appendix \ref{appendix:density}.
Enforcing identical sparsity at every depth suppresses the compute‑hungry layers that naturally benefit from activating more experts, while simultaneously forcing unnecessary activations in layers where sparse representations suffice. Once this bias is lifted, the model is free to self‑organize into a more effective, functionally aligned activation structure.

\begin{table}[t]
\caption{Effect of load balancing interpolation $\mu$. Throughput is measured as samples per second.}
\label{tab:dual_balancing}
\vspace{0.5em}
\centering
\small
\begin{tabular}{rl|cc}
\toprule
& \textbf{$\bm{\mu}$} & \textbf{PPL$\downarrow$} & \textbf{Eval. Throughput $\uparrow$} \\
\midrule
only TB  & 0.0 & 28.41 & 645.7 \\
         & 0.2 & 28.35 & 648.3 \\
balanced & 0.5 & 28.34 & 662.3 \\
         & 0.8 & 28.38 & 643.9 \\
only EB  & 1.0 & 28.43 & 648.8 \\
\bottomrule
\end{tabular}
\end{table}

\begin{figure*}[t]
    \centering
    \begin{subfigure}[b]{0.32\linewidth}
        \centering
        \includegraphics[height=4.5cm]{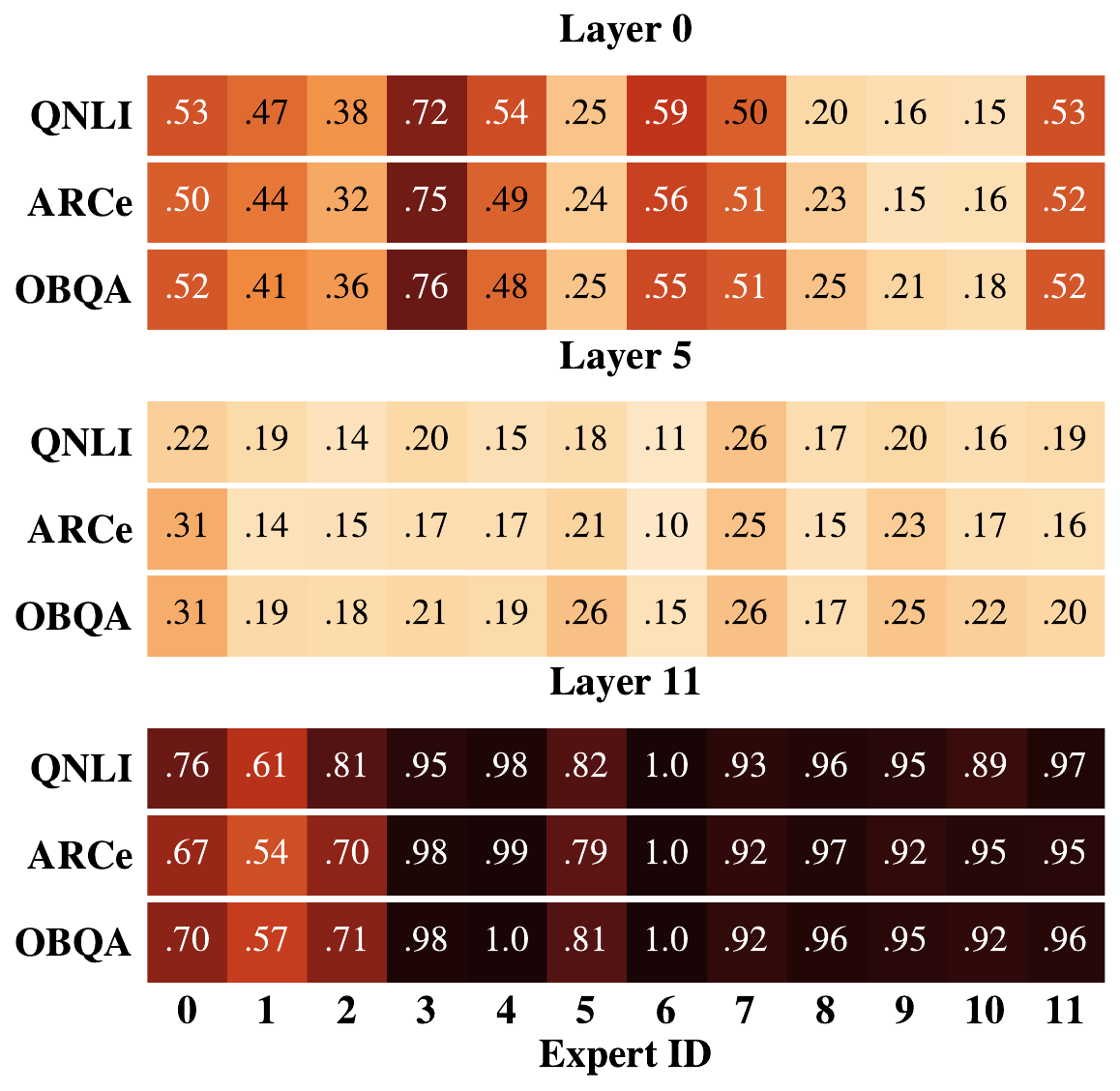}
        \caption{\(\mu=0.0\)}
        \label{fig:token_only}
    \end{subfigure}
    \hfill
    \begin{subfigure}[b]{0.32\linewidth}
        \centering
        \includegraphics[height=4.5cm]{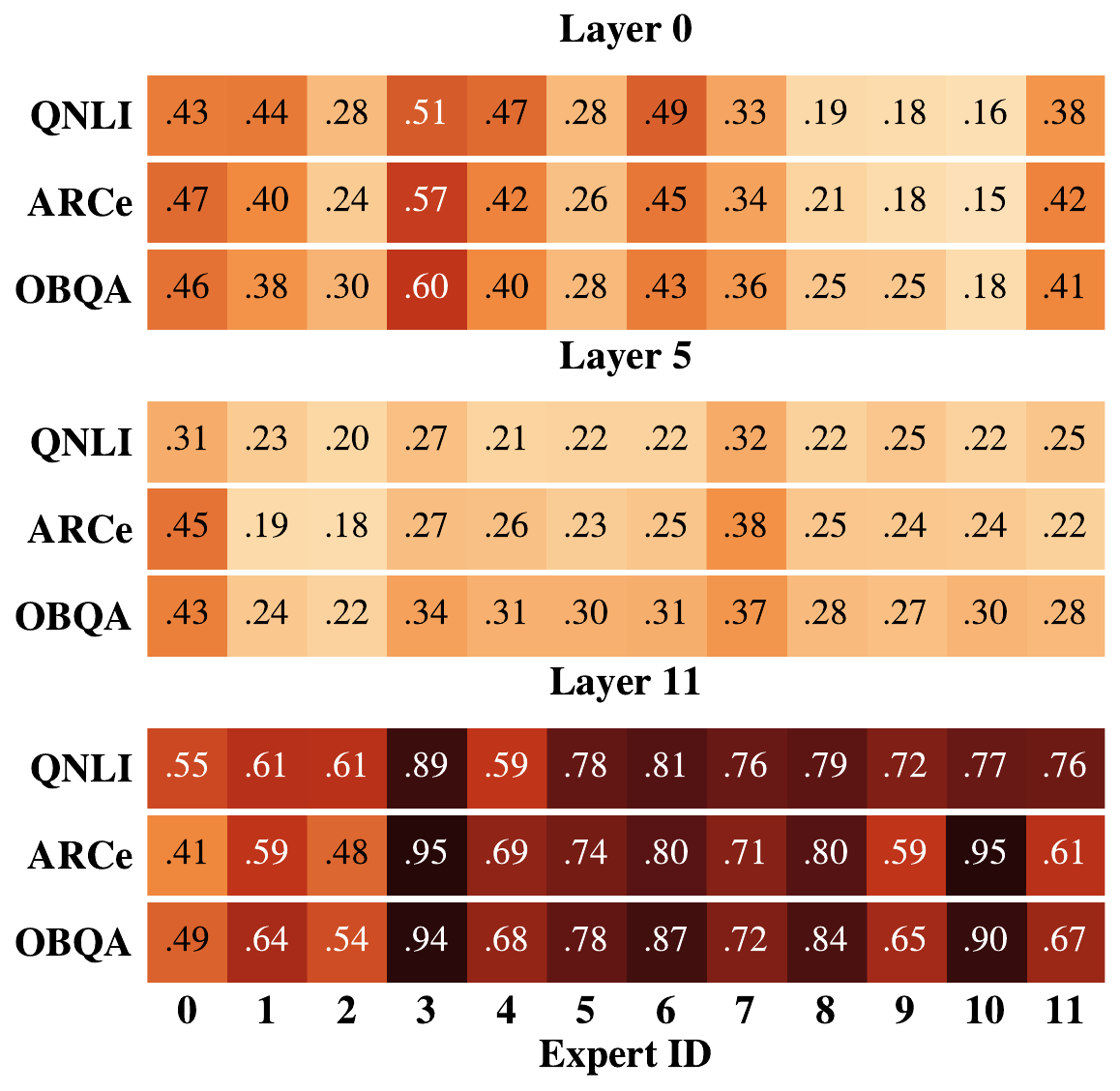}
        \caption{\(\mu=0.5\)}
        \label{fig:dual_banlancing}
    \end{subfigure}
    \hfill
    \begin{subfigure}[b]{0.32\linewidth}
        \centering
        \includegraphics[height=4.5cm]{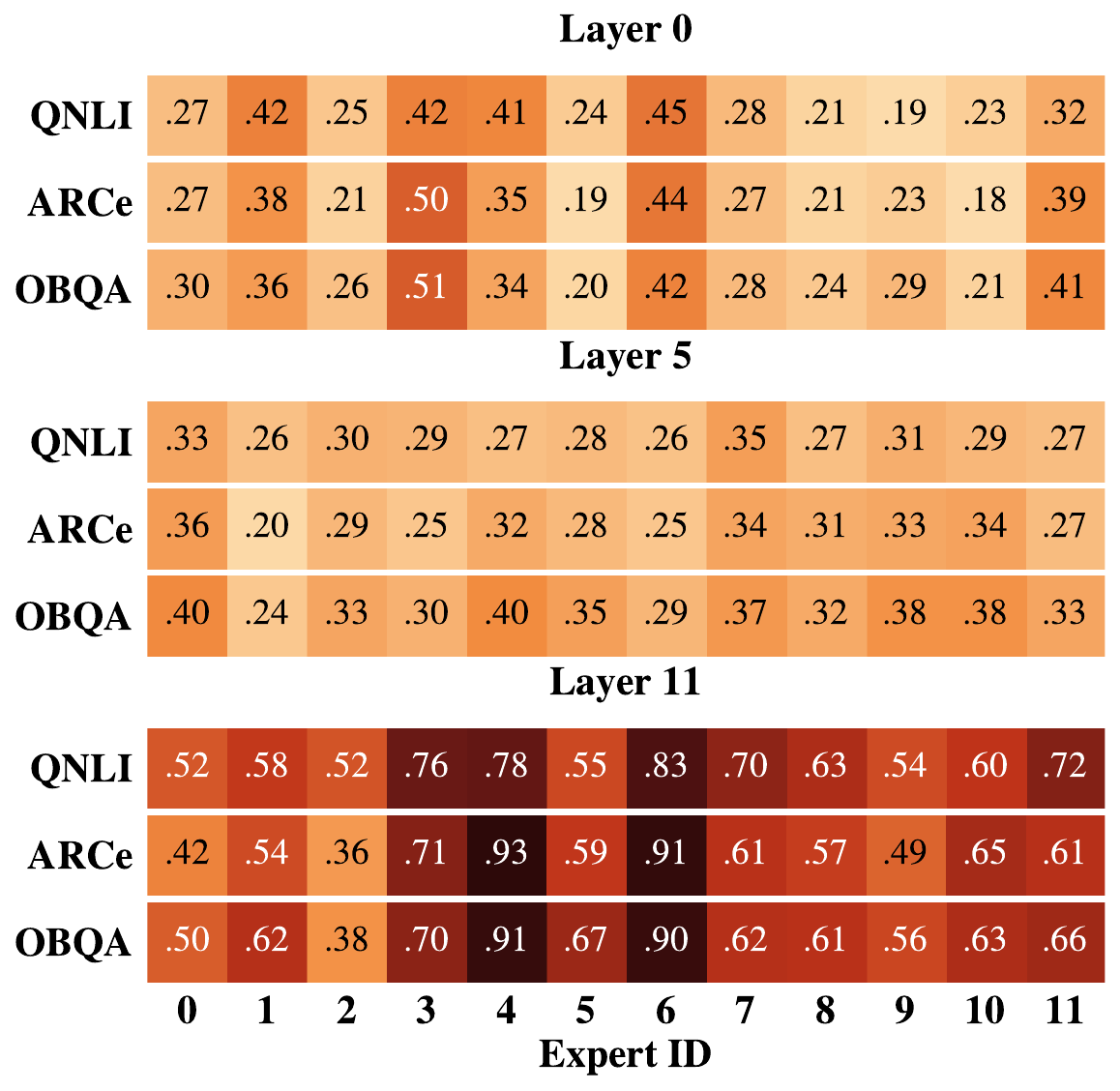}
        \caption{\(\mu=1.0\)}
        \label{fig:expert_only}
    \end{subfigure}
    \caption{Expert activation heatmaps on different input for representative layers at scale S. Each row shows the average expert activation ratios on a given benchmark. Darker colors indicate more frequent activation.}
    \label{fig:TBEB}
    \vspace{-6pt}
\end{figure*}

\subsection{Token and Expert Balancing}

A key contribution of this work is demonstrating empirically that token- and expert-balancing strategies can be combined in a complementary manner via soft interpolation rather than being treated as mutually exclusive. 
Prior work \citep{zhou2022mixture, muennighoff2024olmoe} documented that EC and TC perform better under different configurations, but our unified framework provides a mechanism to adaptively balance both objectives.

Table \ref{tab:dual_balancing} shows that $\mu = 0.5$ achieves the lowest perplexity and highest throughput, which degrades toward either direction.
Token-balancing enables flexible compute allocation based on input characteristics, while expert-balancing ensures uniform expert utilization and encourages specialization; the two objectives address orthogonal failure modes and jointly produce better outcomes than either alone.
This complementarity is illustrated in Figure~\ref{fig:TBEB}. Under token-only balancing (Figure~\ref{fig:token_only}), the absence of expert balancing leads to larger activation probability differences across the expert axis, with a few experts activated far more frequently than others. Conversely, under expert-only balancing (Figure~\ref{fig:expert_only}), expert loads are more horizontally uniform, but activation patterns vary notably across benchmarks, indicating that without token-level regularization, expert activation becomes overly sensitive to the input domain distribution. When both objectives are active ($\mu=0.5$, Figure~\ref{fig:dual_banlancing}), the activation pattern is more uniform along both axes and more consistent across inputs.

\section{Related Work}

\subsection{MoE Foundations}
Mixture-of-Experts was introduced as adaptive mixtures of local experts \citep{jacobs1991adaptive, jordan1994hierarchical} and later scaled to deep networks \citep{eigen2013learning, shazeer2017outrageously}. Interpretability studies reveal that FFNs in transformers capture knowledge with sparse activations \citep{geva2021transformer, dai2022knowledge, dalvi2019one, durrani2020analyzing, gurnee2023finding}, motivating MoE designs that activate only a subset of parameters \citep{liu2023towards}. Modern MoE architectures have since been deployed at scale \citep{lepikhin2020gshard, fedus2022switchtransformer, zoph2022designing, komatsuzaki2022sparse, rajbhandari2022deepspeed, du2022glam}, with frontier models with over billions of parameters \citep{jiang2024mixtral, dai2024deepseekmoe, grok_1}. Structural enhancements such as shared experts \citep{gou2023mixture, dai2024deepseekmoe} further help reduce parameter redundancy.

\subsection{Routing Mechanisms}
Traditional MoE relies on centralized routers, learned linear projections followed by TopK selection \citep{lepikhin2020gshard, fedus2022switchtransformer}. 
Recent work has moved toward relaxing the routing mechanisms. \citet{lv2025aoe} replaces the router with expert-internal scoring; \citet{wang2024remoe} replaces Softmax and TopK with ReLU gating. 
Other approaches include using vector quantization for expert assignment \citep{do2024role}, virtual shared experts \citep{wuunion}, using pretrained language models as routers  \citep{liu2025llm}, and with ternary expert expansion \citep{yan2025tc}. 
\citet{huang2024harder} propose adjusting expert counts based on input difficulty.
\citet{do2025usmoe} utilize global TopK selection over a combined similarity score to unify token- and expert-choice routing.

\rmv{Our work extends the AoE paradigm by removing all centralized constraints (Softmax and TopK), enabling fully decentralized expert self-activation with continuous threshold-based gating.}

\subsection{Load Balancing and Training}
Token Choice \citep{shazeer2017outrageously, fedus2022switchtransformer} guarantees per-token compute but not expert balance.
Expert Choice \citep{zhou2022mixture} guarantees expert load balance, but may potentially yield suboptimal matching.
Studies has shown that Token Choice and Expert Choice balancing are complementary rather than mutually exclusive \citep{muennighoff2024olmoe}. 
Recent advances include auxiliary-loss-free balancing via dynamic bias \citep{wang2024auxiliary, liu2024deepseek}, similarity-preserving routers \citep{omi2025load}, expert specialization through orthogonality losses \citep{guo2025advancing, feng2025dive}, and infrastructure-level scheduling \citep{zhao2025micromoe, yu2025efficient}. For training stability, TopK selection's discontinuity prevents gradient flow to non-selected experts; \citet{wang2024remoe} address this with differentiable ReLU gating and adaptive regularization. \citet{qiu2025demons} provide practical guidance on auxiliary losses, while \citet{he2024expertflow} and \citet{pan2024dense} address auxiliary losses at inference. 
\citet{do2025usmoe} propose a unified scoring function with global TopK selection that linearly combines TC and EC similarity scores, which addresses representation collapse and token dropping simultaneously, but is not applicable without routing mechanisms.
Our approach, instead, seeks unifying token-balancing and expert-balancing under fully decentralized expert self-activation.

\section{Conclusion}
We present Routing-Free Mixture-of-Experts, an MoE architecture that entirely eliminates centralized routing mechanisms, along with a unified adaptive load-balancing framework that jointly optimizes token- and expert-balancing objectives during training. 
Experiments across 3 scales and 9 benchmarks validate that Routing-Free MoE consistently outperforms baselines, with improved scalability and robustness. 
We further analyze the load-balancing behavior throughout training and the expert activation patterns.
Routing-Free MoE opens a path toward more flexible and efficient architectures.
We hope this work encourages future exploration of MoE design and optimization.

\newpage
\section*{Limitations}

Our experiments are conducted exclusively on models trained from scratch on OpenWebText at scales up to 0.8B parameters.
While results are consistent across three scales, it remains an open question whether the findings generalize to even larger-scale pretraining regimes, which lie beyond our current resource limits.
Similarly, we only evaluate on 9 English benchmarks for commonsense reasoning and natural language inference which are still limited in scope and may not capture the full range of applications present in real-world applications. 
Given that our method follows standard training practices common among a broad body of peer-reviewed research in this field, it is highly plausible that the observed trends will carry over to broader settings.

Another practical direction we have not explored is converting an existing pretrained standard MoE model into a Routing-Free MoE via continued pretraining or training-free adaptation, rather than training from scratch.
Such a conversion could possibly further reduce the compute cost of adopting our approach, making it a promising avenue for future work; yet it introduces a vast set of distinct challenges that lies beyond the scope of this work.

\section*{Ethics Consideration}

Routing‑Free MoE is proposed as a general purpose language model architecture and does not introduce domain‑specific risks beyond those inherent to language model pretraining. However, as with any method that improves model efficiency and scalability, it may lower the barrier to training more capable models, with corresponding societal implications that merit careful consideration in downstream applications.

The OpenWebText dataset used in our experiments may also reflect demographic or geographic biases present in large‑scale web corpora. We did not conduct dedicated bias or safety audits, and the absence of such analyses means that potential fairness, privacy, or safety issues may persist.
It is therefore incumbent upon downstream users and deployers to perform appropriate task‑specific fairness, privacy, and safety evaluations before any real‑world deployment. We disclaim responsibility for unintended consequences arising from downstream use that involve applying models trained using our approach.

\rmv{
Our analysis of deployment efficiency under expert parallelism in Appendix~\ref{app:deployment} is based on theoretical estimation.
The deployment efficiency results we report are samples-per-second numbers during evaluation, and we have not provided measurements validating the estimated latency and bandwidth advantages under expert parallelism.
Experimental validation of the deployment efficiency claims on real hardware with multiple devices remains unexplored.
}

\section*{Acknowledgments}

The authors gratefully acknowledge the scientific support and HPC resources provided by the Karlsruhe Institute of Technology National High Performance Computing Center (NHR@KIT) under the NHR projects 22189, 22560, and 24767. 
The HoreKa supercomputer at NHR@KIT is funded by the Ministry of Science, Research and the Arts Baden-W\"urttemberg and by the Federal Ministry of Education and Research of Germany.
This work is also partially supported by funding from Munich Center for Machine Learning (MCML).
We also acknowledge the EuroHPC Joint Undertaking for awarding this project access to the EuroHPC supercomputer LEONARDO under project EHPC-AI-2024A06-060, hosted by CINECA (Italy) and the LEONARDO consortium through a EuroHPC Regular Access call.
In addition, the authors thank Xu He and Alinur Kozhanov from Technical University of Munich for their contributions during the early stages of this work.

\bibliography{references}

\begin{thebibliography}{62}
\providecommand{\natexlab}[1]{#1}

\bibitem[{Ainslie et~al.(2023)Ainslie, Lee-Thorp, De~Jong, Zemlyanskiy, Lebr{\'o}n, and Sanghai}]{ainslie2023gqa}
Joshua Ainslie, James Lee-Thorp, Michiel De~Jong, Yury Zemlyanskiy, Federico Lebr{\'o}n, and Sumit Sanghai. 2023.
\newblock \href {https://aclanthology.org/2023.emnlp-main.298/} {Gqa: Training generalized multi-query transformer models from multi-head checkpoints}.
\newblock In \emph{Proceedings of the 2023 Conference on Empirical Methods in Natural Language Processing}, pages 4895--4901.

\bibitem[{Bisk et~al.(2020)Bisk, Zellers, Le~bras, Gao, and Choi}]{bisk2020piqa}
Yonatan Bisk, Rowan Zellers, Ronan Le~bras, Jianfeng Gao, and Yejin Choi. 2020.
\newblock \href {https://ojs.aaai.org/index.php/AAAI/article/view/6239} {Piqa: Reasoning about physical commonsense in natural language}.
\newblock In \emph{Proceedings of the AAAI conference on artificial intelligence}, volume~34, pages 7432--7439.

\bibitem[{Clark et~al.(2018)Clark, Cowhey, Etzioni, Khot, Sabharwal, Schoenick, and Tafjord}]{clark2018arc}
Peter Clark, Isaac Cowhey, Oren Etzioni, Tushar Khot, Ashish Sabharwal, Carissa Schoenick, and Oyvind Tafjord. 2018.
\newblock \href {https://api.semanticscholar.org/CorpusID:3922816} {Think you have solved question answering? try arc, the ai2 reasoning challenge}.
\newblock \emph{ArXiv}, abs/1803.05457.

\bibitem[{Dai et~al.(2024)Dai, Deng, Zhao, Xu, Gao, Chen, Li, Zeng, Yu, Wu, Xie, Li, Huang, Luo, Ruan, Sui, and Liang}]{dai2024deepseekmoe}
Damai Dai, Chengqi Deng, Chenggang Zhao, R.x. Xu, Huazuo Gao, Deli Chen, Jiashi Li, Wangding Zeng, Xingkai Yu, Y.~Wu, Zhenda Xie, Y.k. Li, Panpan Huang, Fuli Luo, Chong Ruan, Zhifang Sui, and Wenfeng Liang. 2024.
\newblock \href {https://doi.org/10.18653/v1/2024.acl-long.70} {{D}eep{S}eek{M}o{E}: Towards ultimate expert specialization in mixture-of-experts language models}.
\newblock In \emph{Proceedings of the 62nd Annual Meeting of the Association for Computational Linguistics (Volume 1: Long Papers)}, pages 1280--1297, Bangkok, Thailand. Association for Computational Linguistics.

\bibitem[{Dai et~al.(2022)Dai, Dong, Hao, Sui, Chang, and Wei}]{dai2022knowledge}
Damai Dai, Li~Dong, Yaru Hao, Zhifang Sui, Baobao Chang, and Furu Wei. 2022.
\newblock \href {https://aclanthology.org/2022.acl-long.581/} {Knowledge neurons in pretrained transformers}.
\newblock In \emph{Proceedings of the 60th Annual Meeting of the Association for Computational Linguistics (Volume 1: Long Papers)}, pages 8493--8502.

\bibitem[{Dalvi et~al.(2019)Dalvi, Durrani, Sajjad, Belinkov, Bau, and Glass}]{dalvi2019one}
Fahim Dalvi, Nadir Durrani, Hassan Sajjad, Yonatan Belinkov, Anthony Bau, and James Glass. 2019.
\newblock \href {https://ojs.aaai.org/index.php/AAAI/article/view/4592} {What is one grain of sand in the desert? analyzing individual neurons in deep nlp models}.
\newblock In \emph{Proceedings of the AAAI Conference on Artificial Intelligence}, volume~33, pages 6309--6317.

\bibitem[{Dauphin et~al.(2017)Dauphin, Fan, Auli, and Grangier}]{dauphin2017language}
Yann~N Dauphin, Angela Fan, Michael Auli, and David Grangier. 2017.
\newblock \href {http://proceedings.mlr.press/v70/dauphin17a.html} {Language modeling with gated convolutional networks}.
\newblock In \emph{International conference on machine learning}, pages 933--941. PMLR.

\bibitem[{DeepSeek-AI et~al.(2025)DeepSeek-AI, Liu, Feng, Xue, Wang, Wu, Lu, Zhao, Deng, Zhang, Ruan, Dai, Guo, Yang, Chen, Ji, Li, Lin, Dai, Luo, Hao, Chen, Li, Zhang, Bao, Xu, Wang, Zhang, Ding, Xin, Gao, Li, Qu, Cai, Liang, Guo, Ni, Li, Wang, Chen, Chen, Yuan, Qiu, Li, Song, Dong, Hu, Gao, Guan, Huang, Yu, Wang, Zhang, Xu, Xia, Zhao, Wang, Zhang, Li, Wang, Zhang, Zhang, Tang, Li, Tian, Huang, Wang, Zhang, Wang, Zhu, Chen, Du, Chen, Jin, Ge, Zhang, Pan, Wang, Xu, Zhang, Chen, Li, Lu, Zhou, Chen, Wu, Ye, Ye, Ma, Wang, Zhou, Yu, Zhou, Pan, Wang, Yun, Pei, Sun, Xiao, Zeng, Zhao, An, Liu, Liang, Gao, Yu, Zhang, Li, Jin, Wang, Bi, Liu, Wang, Shen, Chen, Zhang, Chen, Nie, Sun, Wang, Cheng, Liu, Xie, Liu, Yu, Song, Shan, Zhou, Yang, Li, Su, Lin, Li, Wang, Wei, Zhu, Zhang, Xu, Xu, Huang, Li, Zhao, Sun, Li, Wang, Yu, Zheng, Zhang, Shi, Xiong, He, Tang, Piao, Wang, Tan, Ma, Liu, Guo, Wu, Ou, Zhu, Wang, Gong, Zou, He, Zha, Xiong, Ma, Yan, Luo, You, Liu, Zhou, Wu, Ren, Ren, Sha, Fu, Xu, Huang, Zhang, Xie, Zhang, Hao,
  Gou, Ma, Yan, Shao, Xu, Wu, Zhang, Li, Gu, Zhu, Liu, Li, Xie, Song, Gao, and Pan}]{liu2024deepseek}
DeepSeek-AI, Aixin Liu, Bei Feng, Bing Xue, Bingxuan Wang, Bochao Wu, Chengda Lu, Chenggang Zhao, Chengqi Deng, Chenyu Zhang, Chong Ruan, Damai Dai, Daya Guo, Dejian Yang, Deli Chen, Dongjie Ji, Erhang Li, Fangyun Lin, Fucong Dai, and 181 others. 2025.
\newblock \href {https://arxiv.org/abs/2412.19437} {Deepseek-v3 technical report}.

\bibitem[{Do et~al.(2025)Do, Le, and Tran}]{do2025usmoe}
Giang Do, Hung Le, and Truyen Tran. 2025.
\newblock \href {https://arxiv.org/abs/2503.22996} {Unified sparse mixture of experts}.
\newblock \emph{arXiv preprint arXiv:2503.22996}.

\bibitem[{Do et~al.(2024)Do, Pham, Le, and Tran}]{do2024role}
Giang Do, Kha Pham, Hung Le, and Truyen Tran. 2024.
\newblock \href {https://arxiv.org/abs/2411.19402} {On the role of discrete representation in sparse mixture of experts}.
\newblock \emph{arXiv preprint arXiv:2411.19402}.

\bibitem[{Du et~al.(2022)Du, Huang, Dai, Tong, Lepikhin, Xu, Krikun, Zhou, Yu, Firat, Zoph, Fedus, Bosma, Zhou, Wang, Wang, Webster, Pellat, Robinson, Meier-Hellstern, Duke, Dixon, Zhang, Le, Wu, Chen, and Cui}]{du2022glam}
Nan Du, Yanping Huang, Andrew~M Dai, Simon Tong, Dmitry Lepikhin, Yuanzhong Xu, Maxim Krikun, Yanqi Zhou, Adams~Wei Yu, Orhan Firat, Barret Zoph, Liam Fedus, Maarten~P Bosma, Zongwei Zhou, Tao Wang, Emma Wang, Kellie Webster, Marie Pellat, Kevin Robinson, and 8 others. 2022.
\newblock \href {https://proceedings.mlr.press/v162/du22c.html} {{GL}a{M}: Efficient scaling of language models with mixture-of-experts}.
\newblock In \emph{Proceedings of the 39th International Conference on Machine Learning}, volume 162 of \emph{Proceedings of Machine Learning Research}, pages 5547--5569. PMLR.

\bibitem[{Durrani et~al.(2020)Durrani, Sajjad, Dalvi, and Belinkov}]{durrani2020analyzing}
Nadir Durrani, Hassan Sajjad, Fahim Dalvi, and Yonatan Belinkov. 2020.
\newblock \href {https://aclanthology.org/2020.emnlp-main.395/} {Analyzing individual neurons in pre-trained language models}.
\newblock In \emph{Proceedings of the 2020 Conference on Empirical Methods in Natural Language Processing (EMNLP)}, pages 4865--4880.

\bibitem[{Eigen et~al.(2013)Eigen, Ranzato, and Sutskever}]{eigen2013learning}
David Eigen, Marc'Aurelio Ranzato, and Ilya Sutskever. 2013.
\newblock \href {https://arxiv.org/abs/1312.4314} {Learning factored representations in a deep mixture of experts}.
\newblock \emph{arXiv preprint arXiv:1312.4314}.

\bibitem[{Fedus et~al.(2022)Fedus, Zoph, and Shazeer}]{fedus2022switchtransformer}
William Fedus, Barret Zoph, and Noam Shazeer. 2022.
\newblock \href {https://www.jmlr.org/papers/v23/21-0998.html} {Switch transformers: Scaling to trillion parameter models with simple and efficient sparsity}.
\newblock \emph{Journal of Machine Learning Research}, 23(120):1--39.

\bibitem[{Feng et~al.(2025)Feng, Shen, Gu, Zhao, Fu, Lin, and Wang}]{feng2025dive}
Yuchen Feng, Bowen Shen, Naibin Gu, Jiaxuan Zhao, Peng Fu, Zheng Lin, and Weiping Wang. 2025.
\newblock \href {https://aclanthology.org/2025.acl-long.951/} {Dive into moe: Diversity-enhanced reconstruction of large language models from dense into mixture-of-experts}.
\newblock In \emph{Proceedings of the 63rd Annual Meeting of the Association for Computational Linguistics (Volume 1: Long Papers)}, pages 19375--19394.

\bibitem[{Gao et~al.(2024{\natexlab{a}})Gao, Chen, Rao, Sun, Liu, Peng, Zhang, Guo, Yang, and Subrahmanian}]{gao2024mola}
Chongyang Gao, Kezhen Chen, Jinmeng Rao, Baochen Sun, Ruibo Liu, Daiyi Peng, Yawen Zhang, Xiaoyuan Guo, Jie Yang, and VS~Subrahmanian. 2024{\natexlab{a}}.
\newblock \href {https://arxiv.org/abs/2402.08562} {Higher layers need more lora experts}.
\newblock \emph{arXiv preprint arXiv:2402.08562}.

\bibitem[{Gao et~al.(2024{\natexlab{b}})Gao, Tow, Abbasi, Biderman, Black, DiPofi, Foster, Golding, Hsu, Le~Noac'h, Li, McDonell, Muennighoff, Ociepa, Phang, Reynolds, Schoelkopf, Skowron, Sutawika, Tang, Thite, Wang, Wang, and Zou}]{gao2021framework}
Leo Gao, Jonathan Tow, Baber Abbasi, Stella Biderman, Sid Black, Anthony DiPofi, Charles Foster, Laurence Golding, Jeffrey Hsu, Alain Le~Noac'h, Haonan Li, Kyle McDonell, Niklas Muennighoff, Chris Ociepa, Jason Phang, Laria Reynolds, Hailey Schoelkopf, Aviya Skowron, Lintang Sutawika, and 5 others. 2024{\natexlab{b}}.
\newblock \href {https://doi.org/10.5281/zenodo.12608602} {The language model evaluation harness}.

\bibitem[{Geva et~al.(2021)Geva, Schuster, Berant, and Levy}]{geva2021transformer}
Mor Geva, Roei Schuster, Jonathan Berant, and Omer Levy. 2021.
\newblock \href {https://aclanthology.org/2021.emnlp-main.446/} {Transformer feed-forward layers are key-value memories}.
\newblock In \emph{Proceedings of the 2021 Conference on Empirical Methods in Natural Language Processing}, pages 5484--5495.

\bibitem[{Gokaslan et~al.(2019)Gokaslan, Cohen, Pavlick, and Tellex}]{Gokaslan2019OpenWeb}
Aaron Gokaslan, Vanya Cohen, Ellie Pavlick, and Stefanie Tellex. 2019.
\newblock Openwebtext corpus.
\newblock \url{http://Skylion007.github.io/OpenWebTextCorpus}.

\bibitem[{Gou et~al.(2023)Gou, Liu, Chen, Hong, Xu, Li, Yeung, Kwok, and Zhang}]{gou2023mixture}
Yunhao Gou, Zhili Liu, Kai Chen, Lanqing Hong, Hang Xu, Aoxue Li, Dit-Yan Yeung, James~T Kwok, and Yu~Zhang. 2023.
\newblock \href {https://arxiv.org/abs/2312.12379} {Mixture of cluster-conditional lora experts for vision-language instruction tuning}.
\newblock \emph{arXiv preprint arXiv:2312.12379}.

\bibitem[{Grok(2024)}]{grok_1}
Grok. 2024.
\newblock \href {https://x.ai/blog/grok-os} {Open release of grok-1}.

\bibitem[{Guo et~al.(2025{\natexlab{a}})Guo, Yang, Zhang, Song, Wang, Zhu, Xu, Zhang, Ma, Bi, Zhang, Yu, Wu, Wu, Gou, Shao, Li, Gao, Liu, Xue, Wang, Wu, Feng, Lu, Zhao, Deng, Ruan, Dai, Chen, Ji, Li, Lin, Dai, Luo, Hao, Chen, Li, Zhang, Xu, Ding, Gao, Qu, Li, Guo, Li, Chen, Yuan, Tu, Qiu, Li, Cai, Ni, Liang, Chen, Dong, Hu, You, Gao, Guan, Huang, Yu, Wang, Zhang, Zhao, Wang, Zhang, Xu, Xia, Zhang, Zhang, Tang, Zhou, Li, Wang, Li, Tian, Huang, Zhang, Wang, Chen, Du, Ge, Zhang, Pan, Wang, Chen, Jin, Chen, Lu, Zhou, Chen, Ye, Wang, Yu, Zhou, Pan, Li, Zhou, Wu, Yun, Pei, Sun, Wang, Zeng, Liu, Liang, Gao, Yu, Zhang, Xiao, An, Liu, Wang, Chen, Nie, Cheng, Liu, Xie, Liu, Yang, Li, Su, Lin, Li, Jin, Shen, Chen, Sun, Wang, Song, Zhou, Wang, Shan, Li, Wang, Wei, Zhang, Xu, Li, Zhao, Sun, Wang, Yu, Zhang, Shi, Xiong, He, Piao, Wang, Tan, Ma, Liu, Guo, Ou, Wang, Gong, Zou, He, Xiong, Luo, You, Liu, Zhou, Zhu, Huang, Li, Zheng, Zhu, Ma, Tang, Zha, Yan, Ren, Ren, Sha, Fu, Xu, Xie, Zhang, Hao, Ma, Yan, Wu, Gu, Zhu, Liu, Li,
  Xie, Song, Pan, Huang, Xu, Zhang, and Zhang}]{guo2025deepseek}
Daya Guo, Dejian Yang, Haowei Zhang, Junxiao Song, Peiyi Wang, Qihao Zhu, Runxin Xu, Ruoyu Zhang, Shirong Ma, Xiao Bi, Xiaokang Zhang, Xingkai Yu, Yu~Wu, ZF~Wu, Zhibin Gou, Zhihong Shao, Zhuoshu Li, Ziyi Gao, Aixin Liu, and 175 others. 2025{\natexlab{a}}.
\newblock \href {https://www.nature.com/articles/s41586-025-09422-z} {Deepseek-r1 incentivizes reasoning in llms through reinforcement learning}.
\newblock \emph{Nature}, 645(8081):633--638.

\bibitem[{Guo et~al.(2025{\natexlab{b}})Guo, Lu, Nan, Chu, Zhuang, Yang, Che, Cao, Leng, Cui, and Jiang}]{guo2025advancing}
Hongcan Guo, Haolang Lu, Guoshun Nan, Bolun Chu, Jialin Zhuang, Yuan Yang, Wenhao Che, Xinye Cao, Sicong Leng, Qimei Cui, and Xudong Jiang. 2025{\natexlab{b}}.
\newblock \href {https://openreview.net/forum?id=iydmH9boLb} {Advancing expert specialization for better moe}.
\newblock In \emph{The Thirty-ninth Annual Conference on Neural Information Processing Systems}.

\bibitem[{Gurnee et~al.(2023)Gurnee, Nanda, Pauly, Harvey, Troitskii, and Bertsimas}]{gurnee2023finding}
Wes Gurnee, Neel Nanda, Matthew Pauly, Katherine Harvey, Dmitrii Troitskii, and Dimitris Bertsimas. 2023.
\newblock \href {https://arxiv.org/abs/2305.01610} {Finding neurons in a haystack: Case studies with sparse probing}.
\newblock \emph{arXiv preprint arXiv:2305.01610}.

\bibitem[{He et~al.(2024)He, Zhang, Wang, Yin, Zeng, Shi, Tang, Chu, Tsang, and Soon}]{he2024expertflow}
Xin He, Shunkang Zhang, Yuxin Wang, Haiyan Yin, Zihao Zeng, Shaohuai Shi, Zhenheng Tang, Xiaowen Chu, Ivor Tsang, and Ong~Yew Soon. 2024.
\newblock \href {https://arxiv.org/abs/2410.17954} {Expertflow: Optimized expert activation and token allocation for efficient mixture-of-experts inference}.
\newblock \emph{arXiv preprint arXiv:2410.17954}.

\bibitem[{Hockney(1994)}]{hockney1994communication}
Roger~W. Hockney. 1994.
\newblock \href {https://doi.org/10.1016/S0167-8191(06)80021-9} {The communication challenge for mpp: Intel paragon and meiko cs-2}.
\newblock \emph{Parallel Comput.}, 20(3):389–398.

\bibitem[{Huang et~al.(2024)Huang, An, Zhuang, Tao, Zhang, Jin, Xu, Chen, Huang, and Feng}]{huang2024harder}
Quzhe Huang, Zhenwei An, Nan Zhuang, Mingxu Tao, Chen Zhang, Yang Jin, Kun Xu, Liwei Chen, Songfang Huang, and Yansong Feng. 2024.
\newblock \href {https://aclanthology.org/2024.acl-long.696/} {Harder task needs more experts: Dynamic routing in moe models}.
\newblock In \emph{Proceedings of the 62nd Annual Meeting of the Association for Computational Linguistics (Volume 1: Long Papers)}, pages 12883--12895.

\bibitem[{Jacobs et~al.(1991)Jacobs, Jordan, Nowlan, and Hinton}]{jacobs1991adaptive}
Robert~A Jacobs, Michael~I Jordan, Steven~J Nowlan, and Geoffrey~E Hinton. 1991.
\newblock \href {https://ieeexplore.ieee.org/abstract/document/6797059} {Adaptive mixtures of local experts}.
\newblock \emph{Neural computation}, 3(1):79--87.

\bibitem[{Jiang et~al.(2024)Jiang, Sablayrolles, Roux, Mensch, Savary, Bamford, Chaplot, de~las Casas, Hanna, Bressand, Lengyel, Bour, Lample, Lavaud, Saulnier, Lachaux, Stock, Subramanian, Yang, Antoniak, Scao, Gervet, Lavril, Wang, Lacroix, and Sayed}]{jiang2024mixtral}
Albert~Q. Jiang, Alexandre Sablayrolles, Antoine Roux, Arthur Mensch, Blanche Savary, Chris Bamford, Devendra~Singh Chaplot, Diego de~las Casas, Emma~Bou Hanna, Florian Bressand, Gianna Lengyel, Guillaume Bour, Guillaume Lample, Lélio~Renard Lavaud, Lucile Saulnier, Marie-Anne Lachaux, Pierre Stock, Sandeep Subramanian, Sophia Yang, and 7 others. 2024.
\newblock \href {https://arxiv.org/abs/2401.04088} {Mixtral of experts}.
\newblock \emph{arXiv preprint arXiv:2401.04088}.

\bibitem[{Jiao et~al.(2024)Jiao, Liu, Tang, Matter, Pfeffer, and Anderson}]{jiao2024spin}
Difan Jiao, Yilun Liu, Zhenwei Tang, Daniel Matter, J{\"u}rgen Pfeffer, and Ashton Anderson. 2024.
\newblock \href {https://aclanthology.org/2024.findings-acl.277} {Spin: Sparsifying and integrating internal neurons in large language models for text classification}.
\newblock In \emph{Findings of the Association for Computational Linguistics: ACL 2024}, pages 4666--4682.

\bibitem[{Jordan and Jacobs(1994)}]{jordan1994hierarchical}
Michael~I Jordan and Robert~A Jacobs. 1994.
\newblock \href {https://ieeexplore.ieee.org/abstract/document/6796382} {Hierarchical mixtures of experts and the em algorithm}.
\newblock \emph{Neural computation}, 6(2):181--214.

\bibitem[{Kaplan et~al.(2020)Kaplan, McCandlish, Henighan, Brown, Chess, Child, Gray, Radford, Wu, and Amodei}]{kaplan2020scaling}
Jared Kaplan, Sam McCandlish, Tom Henighan, Tom~B Brown, Benjamin Chess, Rewon Child, Scott Gray, Alec Radford, Jeffrey Wu, and Dario Amodei. 2020.
\newblock \href {https://arxiv.org/pdf/2001.08361/1000} {Scaling laws for neural language models}.
\newblock \emph{arXiv preprint arXiv:2001.08361}.

\bibitem[{Komatsuzaki et~al.(2022)Komatsuzaki, Puigcerver, Lee-Thorp, Ruiz, Mustafa, Ainslie, Tay, Dehghani, and Houlsby}]{komatsuzaki2022sparse}
Aran Komatsuzaki, Joan Puigcerver, James Lee-Thorp, Carlos~Riquelme Ruiz, Basil Mustafa, Joshua Ainslie, Yi~Tay, Mostafa Dehghani, and Neil Houlsby. 2022.
\newblock \href {https://arxiv.org/abs/2212.05055} {Sparse upcycling: Training mixture-of-experts from dense checkpoints}.
\newblock \emph{arXiv preprint arXiv:2212.05055}.

\bibitem[{Lepikhin et~al.(2020)Lepikhin, Lee, Xu, Chen, Firat, Huang, Krikun, Shazeer, and Chen}]{lepikhin2020gshard}
Dmitry Lepikhin, HyoukJoong Lee, Yuanzhong Xu, Dehao Chen, Orhan Firat, Yanping Huang, Maxim Krikun, Noam Shazeer, and Zhifeng Chen. 2020.
\newblock \href {https://arxiv.org/abs/2006.16668} {Gshard: Scaling giant models with conditional computation and automatic sharding}.
\newblock \emph{arXiv preprint arXiv:2006.16668}.

\bibitem[{Liu and Lo(2025)}]{liu2025llm}
Kuan-Ming Liu and Ming-Chih Lo. 2025.
\newblock \href {https://arxiv.org/abs/2501.09636} {Llm-based routing in mixture of experts: A novel framework for trading}.
\newblock \emph{arXiv preprint arXiv:2501.09636}.

\bibitem[{Liu et~al.(2025)Liu, Ma, Lu, Chen, Ding, and Tresp}]{liu2025parameter}
Yilun Liu, Yunpu Ma, Yuetian Lu, Shuo Chen, Zifeng Ding, and Volker Tresp. 2025.
\newblock \href {https://arxiv.org/abs/2508.02587} {Parameter-efficient routed fine-tuning: Mixture-of-experts demands mixture of adaptation modules}.
\newblock \emph{arXiv preprint arXiv:2508.02587}.

\bibitem[{Liu et~al.(2023)Liu, Dettmers, Lin, Stoyanov, and Li}]{liu2023towards}
Zeyu Liu, Tim Dettmers, Xi~Lin, Veselin Stoyanov, and Xian Li. 2023.
\newblock \href {https://aclanthology.org/2023.emnlp-main.930/} {Towards a unified view of sparse feed-forward network in pretraining large language model}.
\newblock In \emph{Proceedings of the 2023 Conference on Empirical Methods in Natural Language Processing}, pages 15038--15061.

\bibitem[{Lv et~al.(2025)Lv, Xie, Qian, Wu, Sun, Kang, Wang, and Yan}]{lv2025aoe}
Ang Lv, Ruobing Xie, Yining Qian, Songhao Wu, Xingwu Sun, Zhanhui Kang, Di~Wang, and Rui Yan. 2025.
\newblock \href {https://arxiv.org/abs/2501.13074} {Autonomy-of-experts models}.
\newblock \emph{arXiv preprint arXiv:2501.13074}.

\bibitem[{Mihaylov et~al.(2018)Mihaylov, Clark, Khot, and Sabharwal}]{mihaylov2018obqa}
Todor Mihaylov, Peter Clark, Tushar Khot, and Ashish Sabharwal. 2018.
\newblock \href {https://aclanthology.org/D18-1260/} {Can a suit of armor conduct electricity? a new dataset for open book question answering}.
\newblock In \emph{Proceedings of the 2018 conference on empirical methods in natural language processing}, pages 2381--2391.

\bibitem[{Muennighoff et~al.(2025)Muennighoff, Soldaini, Groeneveld, Lo, Morrison, Min, Shi, Walsh, Tafjord, Lambert, Gu, Arora, Bhagia, Schwenk, Wadden, Wettig, Hui, Dettmers, Kiela, Farhadi, Smith, Koh, Singh, and Hajishirzi}]{muennighoff2024olmoe}
Niklas Muennighoff, Luca Soldaini, Dirk Groeneveld, Kyle Lo, Jacob Morrison, Sewon Min, Weijia Shi, Evan~Pete Walsh, Oyvind Tafjord, Nathan Lambert, Yuling Gu, Shane Arora, Akshita Bhagia, Dustin Schwenk, David Wadden, Alexander Wettig, Binyuan Hui, Tim Dettmers, Douwe Kiela, and 5 others. 2025.
\newblock \href {https://openreview.net/forum?id=xXTkbTBmqq} {{OLM}oe: Open mixture-of-experts language models}.
\newblock In \emph{The Thirteenth International Conference on Learning Representations}.

\bibitem[{Omi et~al.(2025)Omi, Sen, and Farhadi}]{omi2025load}
Nabil Omi, Siddhartha Sen, and Ali Farhadi. 2025.
\newblock \href {https://arxiv.org/abs/2506.14038} {Load balancing mixture of experts with similarity preserving routers}.
\newblock \emph{arXiv preprint arXiv:2506.14038}.

\bibitem[{Pan et~al.(2024)Pan, Shen, Liu, Mishra, Zhang, Oliva, Raffel, and Panda}]{pan2024dense}
Bowen Pan, Yikang Shen, Haokun Liu, Mayank Mishra, Gaoyuan Zhang, Aude Oliva, Colin Raffel, and Rameswar Panda. 2024.
\newblock \href {https://arxiv.org/abs/2404.05567} {Dense training, sparse inference: Rethinking training of mixture-of-experts language models}.
\newblock \emph{arXiv preprint arXiv:2404.05567}.

\bibitem[{Qiu et~al.(2025)Qiu, Huang, Zheng, Wen, Wang, Men, Titov, Liu, Zhou, and Lin}]{qiu2025demons}
Zihan Qiu, Zeyu Huang, Bo~Zheng, Kaiyue Wen, Zekun Wang, Rui Men, Ivan Titov, Dayiheng Liu, Jingren Zhou, and Junyang Lin. 2025.
\newblock \href {https://aclanthology.org/2025.acl-long.249/} {Demons in the detail: On implementing load balancing loss for training specialized mixture-of-expert models}.
\newblock In \emph{Proceedings of the 63rd Annual Meeting of the Association for Computational Linguistics (Volume 1: Long Papers)}, pages 5005--5018.

\bibitem[{Rajbhandari et~al.(2022)Rajbhandari, Li, Yao, Zhang, Aminabadi, Awan, Rasley, and He}]{rajbhandari2022deepspeed}
Samyam Rajbhandari, Conglong Li, Zhewei Yao, Minjia Zhang, Reza~Yazdani Aminabadi, Ammar~Ahmad Awan, Jeff Rasley, and Yuxiong He. 2022.
\newblock \href {https://proceedings.mlr.press/v162/rajbhandari22a.html?ref=https://githubhelp.com} {Deepspeed-moe: Advancing mixture-of-experts inference and training to power next-generation ai scale}.
\newblock In \emph{International conference on machine learning}, pages 18332--18346. PMLR.

\bibitem[{Sakaguchi et~al.(2021)Sakaguchi, Bras, Bhagavatula, and Choi}]{sakaguchi2021winogrande}
Keisuke Sakaguchi, Ronan~Le Bras, Chandra Bhagavatula, and Yejin Choi. 2021.
\newblock \href {https://dl.acm.org/doi/10.1145/3474381} {Winogrande: An adversarial winograd schema challenge at scale}.
\newblock \emph{Communications of the ACM}, 64(9):99--106.

\bibitem[{Shazeer(2020)}]{shazeer2020glu}
Noam Shazeer. 2020.
\newblock \href {https://arxiv.org/abs/2002.05202} {Glu variants improve transformer}.
\newblock \emph{arXiv preprint arXiv:2002.05202}.

\bibitem[{Shazeer et~al.(2017)Shazeer, Mirhoseini, Maziarz, Davis, Le, Hinton, and Dean}]{shazeer2017outrageously}
Noam Shazeer, Azalia Mirhoseini, Krzysztof Maziarz, Andy Davis, Quoc Le, Geoffrey Hinton, and Jeff Dean. 2017.
\newblock \href {https://openreview.net/forum?id=B1ckMDqlg} {Outrageously large neural networks: The sparsely-gated mixture-of-experts layer}.
\newblock In \emph{International Conference on Learning Representations}.

\bibitem[{Su et~al.(2024)Su, Ahmed, Lu, Pan, Bo, and Liu}]{su2024roformer}
Jianlin Su, Murtadha Ahmed, Yu~Lu, Shengfeng Pan, Wen Bo, and Yunfeng Liu. 2024.
\newblock \href {https://dl.acm.org/doi/10.1016/j.neucom.2023.127063} {Roformer: Enhanced transformer with rotary position embedding}.
\newblock \emph{Neurocomputing}, 568:127063.

\bibitem[{Vaswani et~al.(2017)Vaswani, Shazeer, Parmar, Uszkoreit, Jones, Gomez, Kaiser, and Polosukhin}]{vaswani2017attention}
Ashish Vaswani, Noam Shazeer, Niki Parmar, Jakob Uszkoreit, Llion Jones, Aidan~N Gomez, {\L}ukasz Kaiser, and Illia Polosukhin. 2017.
\newblock \href {https://proceedings.neurips.cc/paper/2017/hash/3f5ee243547dee91fbd053c1c4a845aa-Abstract.html} {Attention is all you need}.
\newblock \emph{Advances in neural information processing systems}, 30.

\bibitem[{Wang et~al.(2018)Wang, Singh, Michael, Hill, Levy, and Bowman}]{wang2018glue}
Alex Wang, Amanpreet Singh, Julian Michael, Felix Hill, Omer Levy, and Samuel Bowman. 2018.
\newblock \href {https://aclanthology.org/W18-5446/} {Glue: A multi-task benchmark and analysis platform for natural language understanding}.
\newblock In \emph{Proceedings of the 2018 EMNLP workshop BlackboxNLP: Analyzing and interpreting neural networks for NLP}, pages 353--355.

\bibitem[{Wang et~al.(2024{\natexlab{a}})Wang, Gao, Zhao, Sun, and Dai}]{wang2024auxiliary}
Lean Wang, Huazuo Gao, Chenggang Zhao, Xu~Sun, and Damai Dai. 2024{\natexlab{a}}.
\newblock \href {https://arxiv.org/abs/2408.15664} {Auxiliary-loss-free load balancing strategy for mixture-of-experts}.
\newblock \emph{arXiv preprint arXiv:2408.15664}.

\bibitem[{Wang et~al.(2024{\natexlab{b}})Wang, Zhu, and Chen}]{wang2024remoe}
Ziteng Wang, Jun Zhu, and Jianfei Chen. 2024{\natexlab{b}}.
\newblock \href {https://openreview.net/forum?id=4D0f16Vwc3} {Remoe: Fully differentiable mixture-of-experts with relu routing}.
\newblock In \emph{The Thirteenth International Conference on Learning Representations}.

\bibitem[{Wendler et~al.(2024)Wendler, Veselovsky, Monea, and West}]{wendler2024llamas}
Chris Wendler, Veniamin Veselovsky, Giovanni Monea, and Robert West. 2024.
\newblock \href {https://aclanthology.org/2024.acl-long.820/} {Do llamas work in english? on the latent language of multilingual transformers}.
\newblock In \emph{Proceedings of the 62nd Annual Meeting of the Association for Computational Linguistics (Volume 1: Long Papers)}, pages 15366--15394.

\bibitem[{Wu et~al.()Wu, Lv, Xie, Sun, Wang, Yan, and Lin}]{wuunion}
Songhao Wu, Ang Lv, Ruobing Xie, Xingwu Sun, Di~Wang, Rui Yan, and Yankai Lin.
\newblock \href {https://openreview.net/forum?id=Ksgiup7ZNZ} {Union-of-experts: Experts in mixture-of-experts are secretly routers}.

\bibitem[{Xiong et~al.(2020)Xiong, Yang, He, Zheng, Zheng, Xing, Zhang, Lan, Wang, and Liu}]{xiong2020layer}
Ruibin Xiong, Yunchang Yang, Di~He, Kai Zheng, Shuxin Zheng, Chen Xing, Huishuai Zhang, Yanyan Lan, Liwei Wang, and Tieyan Liu. 2020.
\newblock \href {https://proceedings.mlr.press/v119/xiong20b/xiong20b.pdf} {On layer normalization in the transformer architecture}.
\newblock In \emph{International conference on machine learning}, pages 10524--10533. PMLR.

\bibitem[{Yan et~al.(2025)Yan, Bin, Zhang, Wang, and Lin}]{yan2025tc}
Shen Yan, Xingyan Bin, Sijun Zhang, Yisen Wang, and Zhouchen Lin. 2025.
\newblock \href {https://openreview.net/forum?id=dsP91M4hDL} {Tc-moe: Augmenting mixture of experts with ternary expert choice}.
\newblock In \emph{The Thirteenth International Conference on Learning Representations}.

\bibitem[{Yu et~al.(2025)Yu, Ma, Agarwal, Oswald, Huang, Linsenmaier, Mei, Zhao, Borkar, Rouhani, Nellans, Krashinsky, and Khandelwal}]{yu2025efficient}
Yanpeng Yu, Haiyue Ma, Krish Agarwal, Nicolai Oswald, Qijing Huang, Hugo Linsenmaier, Chunhui Mei, Ritchie Zhao, Ritika Borkar, Bita~Darvish Rouhani, David Nellans, Ronny Krashinsky, and Anurag Khandelwal. 2025.
\newblock \href {https://arxiv.org/abs/2512.09277} {Efficient moe serving in the memory-bound regime: Balance activated experts, not tokens}.
\newblock \emph{arXiv preprint arXiv:2512.09277}.

\bibitem[{Zellers et~al.(2019)Zellers, Holtzman, Bisk, Farhadi, and Choi}]{zellers2019hellaswag}
Rowan Zellers, Ari Holtzman, Yonatan Bisk, Ali Farhadi, and Yejin Choi. 2019.
\newblock \href {https://aclanthology.org/P19-1472.pdf} {Hellaswag: Can a machine really finish your sentence?}
\newblock In \emph{Proceedings of the 57th annual meeting of the association for computational linguistics}, pages 4791--4800.

\bibitem[{Zhang and Sennrich(2019)}]{zhang2019root}
Biao Zhang and Rico Sennrich. 2019.
\newblock \href {https://proceedings.neurips.cc/paper/2019/hash/1e8a19426224ca89e83cef47f1e7f53b-Abstract.html} {Root mean square layer normalization}.
\newblock \emph{Advances in neural information processing systems}, 32.

\bibitem[{Zhao et~al.(2025)Zhao, Wu, Song, and Xu}]{zhao2025micromoe}
Chenqi Zhao, Wenfei Wu, Linhai Song, and Yuchen Xu. 2025.
\newblock \href {https://arxiv.org/abs/2511.16947} {Micromoe: Fine-grained load balancing for mixture-of-experts with token scheduling}.
\newblock \emph{arXiv preprint arXiv:2511.16947}.

\bibitem[{Zhou et~al.(2022)Zhou, Lei, Liu, Du, Huang, Zhao, Dai, Chen, Le, and Laudon}]{zhou2022mixture}
Yanqi Zhou, Tao Lei, Hanxiao Liu, Nan Du, Yanping Huang, Vincent~Y Zhao, Andrew Dai, Zhifeng Chen, Quoc Le, and James Laudon. 2022.
\newblock \href {https://proceedings.neurips.cc/paper_files/paper/2022/hash/2f00ecd787b432c1d36f3de9800728eb-Abstract-Conference.html} {Mixture-of-experts with expert choice routing}.
\newblock \emph{Advances in Neural Information Processing Systems}, 35:7103--7114.

\bibitem[{Zoph(2022)}]{zoph2022designing}
Barret Zoph. 2022.
\newblock \href {https://ieeexplore.ieee.org/abstract/document/9835248} {Designing effective sparse expert models}.
\newblock In \emph{2022 IEEE International Parallel and Distributed Processing Symposium Workshops (IPDPSW)}, pages 1044--1044. IEEE.

\end{thebibliography}

\appendix

\newpage
\appendix
\section{Load-Balancing Losses}
\label{app:lb_theory}

\citet{fedus2022switchtransformer} first introduced a differentiable auxiliary load-balancing loss to MoE, defined as the scaled dot-product 
\begin{align}
    \mathcal{L}_\mathrm{LB}=\alpha \cdot N \cdot \sum_{i=1}^N f_i \cdot P_i,
\end{align}
where $f_i$ is the fraction of tokens dispatched to expert $i$ and $P_i$ is the 
fraction of router probability allocated to expert $i$. Since $f_i$ is non-differentiable, 
the gradient flows through $P_i$ alone, while $f_i$ serves as a fixed per-step weight. 
The loss is minimized under uniform routing, as $\sum_i f_i P_i = 1/N$ when both 
vectors are uniform. Our $\mathcal{L}_\text{EB}$ and $\mathcal{L}_\text{TB}$ follow 
the same product-of-averages structure and gradient strategy, extending it to 
simultaneously encourage uniformity along both the expert and token axes without 
any centralized routing mechanism.
We show that $\mathcal{L}_\text{EB}$ and $\mathcal{L}_\text{TB}$ are minimized 
towards the desired uniform activation state at $\rho_\infty$. Let 
\begin{align}
f_i = \frac{1}{|\mathcal{B}|} 
\sum_{\mathbf{x} \in \mathcal{B}} f_i(\mathbf{x}),\quad
\tilde{g}_i = 
\frac{1}{|\mathcal{B}|} \sum_{\mathbf{x} \in \mathcal{B}} G_i(\mathbf{x})
\end{align}
denote the per-expert mean binary activation and its differentiable proxy. Under the 
constraint that the adaptive coefficient $\lambda_t$ pins the mean density 
$\frac{1}{|\mathcal{E}|}\sum_i f_i = \rho_\infty$, the sum $\sum_i f_i \tilde{g}_i$ 
is minimized when all terms are equal by the rearrangement inequality, achieved 
precisely at $f_i = \tilde{g}_i = \rho_\infty$ for all $i$. While the losses are 
not written as explicit variance terms, any imbalance where over-activated experts 
co-occur with larger proxy values increases $\sum_i f_i \tilde{g}_i$ above the 
baseline $\rho_\infty^2$ attained under perfect balance, since $f_i$ and $\tilde{g}_i$ 
share a common dependence on $G_i(\mathbf{x})$. The excess above $\rho_\infty^2$ 
is therefore a monotone function of the covariance between binary activations and 
their proxies, providing implicit variance penalization through a fully differentiable 
surrogate without requiring additional normalization terms. An identical argument 
applies to $\mathcal{L}_\text{TB}$ by symmetry over the token axis.

\section{Routing-Free MoE at Deployment}
\label{app:deployment}
\subsection{Expert Parallelism}
Deploying MoE at scale often benefits from partitioning experts across multiple devices under \emph{expert parallelism} (EP)~\citep{lepikhin2020gshard}.
In this setting, we trace the full critical path of both architectures and analyze per-regime efficiency.

Consider a single MoE layer distributed to $M$ devices, each hosting $N/M$ experts, and processing a batch of $T$ tokens
with hidden dimension $D$ and $b$ bytes per element. 
We adopt the standard $\alpha$-$\beta$ model~\citep{hockney1994communication} that transferring $n$ bytes over one hop incurs latency $\alpha + n/B$, where
$\alpha$ is the startup per-hop latency and $B$ is the per-link bandwidth.
In standard MoE each token is routed to $K$ experts, and Routing-Free MoE
activates an expected $K_\mathrm{eff} = \rho_\infty \cdot N$ experts per token.

For a standard MoE layer, the router matmul, Softmax, and TopK  are strictly
sequential and centralized, with costs
\begin{align}
    t_\mathrm{routing} & = t_\mathrm{router} + t_\mathrm{Softmax}+ t_\mathrm{TopK}  \notag \\
    &\propto T \cdot (D + 2) \cdot N.
\end{align}
Softmax requires all $N$ logits before any normalization can proceed; TopK  requires a full selection pass over $N$ values. Neither can be parallelized across devices, and dispatch cannot begin until both complete.
Afterwards it produces a token-to-expert index tensor as dispatch assignment used to pack token buffers, which are sent to multiple target devices via blocking All-to-All:
\begin{equation}
    t_\mathrm{A2A}
    = (M{-}1)\,\alpha + \frac{K\cdot T \cdot D \cdot b}{M \cdot B}.
    \label{eq:std-dispatch}
\end{equation}
Expert FFNs run on the received $K \cdot T/N$ tokens per expert with no inter-device communication, with the parallelized per-device cost
\begin{align}
    t_\mathrm{expert} &\propto 3 K \cdot \frac{T}{M} \cdot D \cdot D_\mathrm{act}.
\end{align}
Their outputs are then scattered back to originating devices with the same buffer geometry and All-to-All $t_\mathrm{A2A}$,
and the receiving device uses its locally-stored assignment mask to scatter outputs into the correct positions, which is a cheap local memory operation. Total per-layer cost is
\begin{align}
    T_\mathrm{MoE}
    &= t_\mathrm{routing} + t_\mathrm{expert} \notag \\
    &+2(M{-}1)\,\alpha + 2\; \frac{K\cdot T \cdot D \cdot b}{M \cdot B}.
    \label{eq:std-total}
\end{align}

Routing-Free MoE starts from a non-blocking All-Gather to broadcast the full token batch:
\begin{equation}
    t_\mathrm{AG}
    = (M{-}1)\,\alpha + \frac{(M{-}1) \cdot T \cdot D \cdot b}{M \cdot B}.
\end{equation}
Each device can immediately begin scoring its $N/M$ local experts on incoming chunks, pipelining scoring with communication. 
The total scoring cost per device, including applying $\mathbf{xA}_\mathrm{gate}$, norm, bias, activation and thresholding, is
\begin{equation}
    t_\mathrm{scoring} \propto T \cdot D \cdot r \cdot \frac{N}{M},
\end{equation}
For activated token--expert pairs, $\mathbf{x}\mathbf{A}_{\mathrm{gate},i}$
computed during scoring is directly reused in the FFN forward pass, incurring
zero marginal cost. For non-activated tokens computation terminates after the
rank-$r$ projection, costing only $T \cdot D \cdot r$ rather than a full FFN
pass. The remaining per-device FFN cost for activated pairs is
\begin{equation}
    t_\mathrm{expert}^* \propto K_\mathrm{eff} \cdot \frac{T}{M} \cdot (r+2D) \cdot D_\mathrm{act}.
\end{equation}
Activated outputs are immediately returned as asynchronous point-to-point messages upon completion, with no barrier required:
\begin{equation}
    t_\mathrm{combine}
    = \alpha + \frac{K_\mathrm{eff} \cdot T \cdot D \cdot b}{M \cdot B}.
\end{equation}
Receiving devices accumulate partial sums as results arrive. Total per-layer cost is
\begin{align}
    T_\mathrm{RFMoE}
    &= t_\mathrm{scoring} + t_\mathrm{expert}^*+ M\alpha  \notag \\
    &+ \frac{(M-1+K_\mathrm{eff}) \cdot T \cdot D \cdot b}{M \cdot B}.
    \label{eq:rf-total}
\end{align}

Setting $K = K_\mathrm{eff}$ for an iso-compute comparison, the ratio of computation cost for routing and expert processing becomes
\begin{equation}
    \frac{t_\mathrm{scoring}+t_\mathrm{expert}^*}{t_\mathrm{routing} + t_\mathrm{expert}}=
    \frac{rD+\frac{K}{N}(r+2D)D_\mathrm{act}}{(D+2)M+\frac{K}{N}(3D)D_\mathrm{act}}.
\end{equation}
Since $r \ll D$ and $M\ll D_\textrm{act}$, this ratio is strictly less than $1$, and gets smaller when $M$ grows. 
Therefore, under expert parallelism, Routing-Free MoE always requires less computation per layer than standard MoE.

For communication, the barrier synchronization saving $\Delta\alpha = (M{-}2)\,\alpha$ is
strictly positive for all $M \geq 3$, independent of input size or activation ratio. 
The bandwidth terms differ as
\begin{equation}
    \Delta_B = \frac{(K+1-M)\cdot T \cdot D \cdot b}{M \cdot B},
\end{equation}
which favors Routing-Free MoE when $M < K + 1$.
In the prefill stage where $T$ is large, the $\Delta_B$ term dominates the communication overhead, and Routing-Free MoE is advantageous when the number of devices $M$ is no greater than the number of activated experts $K$ for each layer, which typically holds in practice for advanced large-scale MoE models \citep{guo2025deepseek} under expert-parallel setting at inference. 
In the decode stage where $T{=}1$, the bandwidth term becomes negligible and the critical path is instead dominated by $\alpha$ and sequential computational bottlenecks, and Routing-Free MoE becomes exceptionally well-suited for such high-throughput, low-latency streaming inference scenarios.

To corroborate the theoretical analysis, we evaluate Routing-Free MoE against standard MoE under expert-parallel deployment across varying sequence lengths $T$, device counts $M$, and expert activations $K$ for our trained model at scale S in production-representative settings.
We isolate the prefill and the decode stages. For each configuration, we measured end-to-end  processing time, breaking down contributions from computation, communication, and synchronization overhead.

\begin{table}[t]
\caption{Prefill and autoregressive decode tokens-per-second throughput per device under expert parallelism.
Prefill process $T$=1,024 tokens in a single forward pass; decode generates 128 tokens autoregressively with $T$=1 per step. $K$=3 denotes TopK for standard MoE and $K_\text{eff}$=3 for Routing-Free MoE.
All runs use batch size 1 with input tokens sampled randomly from OpenWebText, bfloat16 precision, and report the mean over 10 repetitions after 3 warmup ones.}
\label{tab:inference-efficiency}
\vspace{0.8em}
\centering
\small
\begin{tabular}{r|ccc|c}
\toprule
\textbf{Setting} & \textbf{$\bm{T}$} & \textbf{$\bm{M}$} & \textbf{$\bm{K}$} & \textbf{Throughput $\uparrow$} \\
\midrule
\multicolumn{5}{c}{\textit{Prefill Stage}} \\
\midrule
MoE & 1,024 & 1  & 3 & 22,346.66 \\
MoE & 1,024 & 2  & 3 & 18,034.84 \\
MoE & 1,024 & 3  & 3 & 18,558.08 \\
MoE & 1,024 & 4  & 3 & 18,355.43 \\
\midrule
RFMoE & 1,024 & 1  & 3 & 22,277.77 \\
RFMoE & 1,024 & 2  & 3 & 22,268.24 \\
RFMoE & 1,024 & 3  & 3 & 21,822.29 \\
RFMoE & 1,024 & 4  & 3 & 21,784.77 \\
\midrule
\multicolumn{5}{c}{\textit{Decode Stage}} \\
\midrule
MoE & 1 & 1  & 3 & 57.90 \\
MoE & 1 & 2  & 3 & 45.65 \\
MoE & 1 & 3  & 3 & 45.61 \\
MoE & 1 & 4  & 3 & 44.67 \\
\midrule
RFMoE & 1 & 1  & 3 & 52.14 \\
RFMoE & 1 & 2  & 3 & 52.05 \\
RFMoE & 1 & 3  & 3 & 50.11 \\
RFMoE & 1 & 4  & 3 & 49.84 \\
\bottomrule
\end{tabular}
\end{table}

Table~\ref{tab:inference-efficiency} reports throughput under expert parallelism for models at scale S. 
Standard MoE throughput drops sharply when moving from $M{=}1$ to $M{=}2$, with Prefill falls by ${\sim}19\%$, and decode
falls by ${\sim}21\%$, with little recovery at $M{=}3$ or $M{=}4$. This reflects the blocking All-to-All dispatches and the sequential, non-parallelizable routing pipeline that must complete before any expert computation begins. 
In contrast, Routing-Free MoE retains ${\geq}97.8\%$ of its single-device prefill throughput and ${\geq}95.6\%$ of its decode throughput at $M{=}4$, consistent with the asynchronous scoring and point-to-point communication pattern described above.
Although Routing-Free MoE is slightly slower than standard MoE on a single device due to the rank-$r$ gating projection applied to all $N$ experts rather than a single $D{\times}N$ router matmul, the crossover occurs already at $M{=}2$, where Routing-Free MoE surpasses MoE by $23\%$ in prefill and $14\%$ in decode throughput. The advantage grows with $M$. At $M{=}4$, Routing-Free MoE delivers $1.19{\times}$ higher prefill throughput and $1.12{\times}$ higher decode throughput. This confirms the theoretical prediction that the communication and synchronization savings of Routing-Free MoE compound with device count, while its computational cost ratio remains strictly below unity and decreases monotonically in $M$.
The decode stage exhibits particularly graceful scaling because the per-step bandwidth term $T \cdot D \cdot b / \left(M \cdot B\right)$ is negligible at
$T{=}1$, leaving the critical path dominated by startup latency $\alpha$ and sequential compute. 
Here, Routing-Free MoE's elimination of the centralized routing barrier translates almost entirely into wall-clock savings, highlighting its performance for latency-sensitive autoregressive serving.

\subsection{Threshold Adaptation
}

The design of global post-activation threshold $\theta$ provides a lightweight, inference-time mechanism for trading computation against model quality, beyond its role in controlling overall sparsity during training.
The routing-free training dynamics naturally encourage each expert to commit decisively to its activation state, which pushes its internal score either well above or well below $\theta$ rather 
than hovering near the boundary. 
As a result, moderate perturbations to $\theta$ at inference
time displace only low-confidence, marginally contributing activations, conferring inherent robustness to threshold miscalibration.

\begin{table}[t]
\caption{Effect of the global threshold $\theta$ on downstream benchmark
average (Avg.) and estimated FLOPs at scale~S.
$\rho_\text{eff}$ denotes the empirical mean activation density at~$\theta$.
$^\dagger$Averages at $\theta \geq 1.5$ are driven by the anomalous SST-2 and OBQA spikes (see Figure~\ref{fig:theta_breakdown}).
}
\label{tab:theta_sweep}
\vspace{0.8em}
\centering
\small
\begin{tabular}{r|rc|l}
\toprule
$\bm{\theta}$ & $\bm{\rho}_\text{eff}$\ \textbf{(\%)} &\textbf{FLOPs} & \textbf{Avg.}\ $\uparrow$ \\
\midrule
0.1 & 100.0 & 120.46M & 39.41\\
0.5 & 94.8 & 118.41M & 39.12\\
0.8 & 61.4 & 105.40M & 39.55  \\
0.9 & 48.9 & 100.32M & 39.73  \\
1.0 & 37.8 & 96.21M & 39.80  \\
1.1 & 29.3 & 92.97M & 39.98  \\
1.2 & 22.9 & 90.37M & 39.77  \\
1.5 & 11.1 & 85.80M & 40.77$^\dagger$ \\
1.9 & 4.3 & 83.14M & 40.23$^\dagger$ \\
\bottomrule
\end{tabular}
\vspace{0.8em}
\end{table}

\begin{figure}[t]
    \centering
    \includegraphics[width=0.5\linewidth]{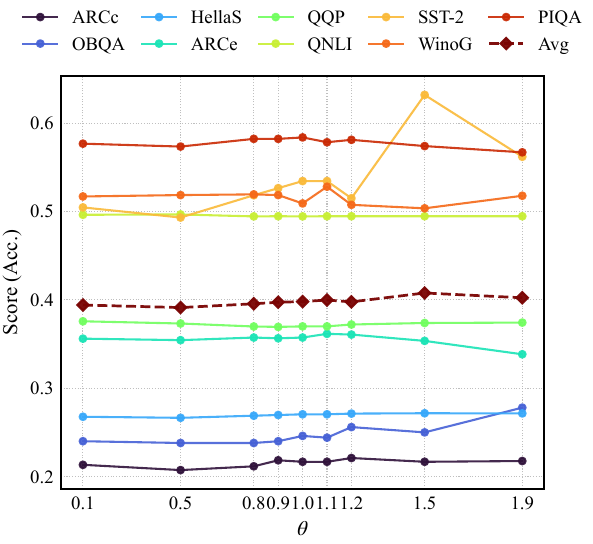}
    \caption{Per-benchmark accuracy across $\theta$ at
    scale~S.  HellaSwag, QQP, QNLI are nearly invariant to the threshold, while PIQA, ARC-easy, ARC-challenge and Winogrande achieve their best performance around $\rho_\text{eff}\approx\rho_\infty$.
    SST-2 and OBQA spikes sharply at larger $\theta$,
    driving the elevated averages.}
    \label{fig:theta_breakdown}
\end{figure}

Table~\ref{tab:theta_sweep} reports the downstream benchmark average across a sweep of $\theta$ at scale~S.  The most striking observation is
the stability of performance despite changes in activation density.  Reducing $\rho_\text{eff}$ from 100\% to
4.3\%, a more than 20$\times$ reduction in expert
activations and a 31\% drop in FLOPs, degrades the average score by less than two absolute points.  
This robustness is a direct consequence of the decisive activation patterns learned during routing-free training.  
Because most expert scores lie far from the decision boundary,
sweeping $\theta$ over a wide range affects only a small number of uncertain, low-impact activations.
Notably, even at extremely low thresholds where nearly all experts are activated, as seen at $\rho_\text{eff} \approx 100\%$ with $\theta{=}0.1$, the model does not suffer from  gradient explosion or output instability.
Since the activation scores $G_i(\mathbf{x})$ directly scale each expert's output before aggregation, experts that pass the threshold with low scores contribute proportionally little to the final representation.
Lowering $\theta$ therefore admits only marginal, low-weight activations rather than introducing equally weighted noise from all experts.

The per-benchmark breakdown in Figure~\ref{fig:theta_breakdown} reveals
that this aggregate stability is not an artifact of compensating trends.
The majority of benchmarks, including HellaSwag, QQP, and QNLI are effectively invariant to $\theta$, confirming that
the core representations remain intact even under aggressive
sparsification.
A small number of tasks exhibit larger sensitivity.
SST-2 in particular spikes at $\theta{=}1.5$, which accounts for most of the elevated aggregate averages at high thresholds.
We attribute this to the interaction between task-specific input distributions and the activation patterns of individual experts, rather than a systematic benefit of sparser computation.
These results highlight a practical advantage of the
routing-free design. The threshold $\theta$ serves as a single, transparent knob that allows practitioners to balance accuracy and efficiency at deployment time without retraining, with assurance that moderate adjustments will not disrupt the expert specialization patterns learned during training.

\section{Statistical Significance Analysis}
\label{app:stat_significance}

To support the claim that \modelname{} consistently outperforms the standard MoE baseline, we conduct a formal statistical analysis over the benchmark results reported in Table~\ref{tab:main_result}.
Each model variant is evaluated on nine benchmarks across three scales, yielding $9 \times 3 = 27$ paired observations, where each pair consists of the accuracy score of \modelname{} and MoE on the same benchmark at the same scale.

We treat each (scale, benchmark) pair as a matched observation.
For each pair we compute the signed improvement $\Delta_i = \text{RFMoE}_i - \text{MoE}_i$ in percentage points (pp).
We test the one-sided null hypothesis $H_0\colon \mu_\Delta \leq 0$ against $H_1\colon \mu_\Delta > 0$.
We apply a one-sided paired $t$-test, which evaluates whether the mean improvement is significantly positive under an approximate normality assumption.
Effect size is quantified via Cohen's $d$ computed on the paired differences, and win rate is reported as a descriptive statistic.

\begin{table}[t]
\caption{Paired statistical tests comparing \modelname{} against the standard MoE baseline across 9 benchmarks. $p$-values are one-sided. }
\label{tab:stat_per_scale}
\vspace{0.8em}
\centering
\small
    \addtolength{\tabcolsep}{-3.4pt} 
\addtolength{\tabcolsep}{2pt}
\begin{tabular}{c|ccccc}
\toprule
\textbf{Scale} & \textbf{W/L} & \textbf{$\Delta$\ Avg.} & \textbf{$\bm{t}$-stat} & \textbf{$\bm{p_t}$} & \textbf{Cohen's $\bm{d}$} \\
\midrule
S  & 5 / 4 & +0.80 & 1.481 & 0.089 & 0.494 \\
M & 6 / 3 & +0.76 & 0.764 & 0.233 & 0.255 \\
L  & 5 / 4 & +0.76 & 1.184 & 0.135 & 0.395  \\
\midrule
Overall & 16 / 11 & +0.77 & 1.858 & 0.037 & 0.358 \\
\bottomrule
\end{tabular}
\end{table}

Table~\ref{tab:stat_per_scale} reports per-scale statistics.
The mean improvement is positive at every scale (+0.80\,pp at S, +0.76\,pp at both M and L), and Cohen's $d$ ranges from 0.26 to 0.49, indicating a consistent small-to-medium effect.
When observations are pooled across all three scales ($n = 27$), the one-sided paired $t$-test yields $t = 1.858$, $p = 0.037$, rejecting $H_0$ at the $\alpha = 0.05$ level.
\modelname{} achieves a positive improvement in 16 out of 27 pairs, with a mean gain of $+0.77$\,pp and Cohen's $d = 0.36$.
Beyond downstream task accuracy, \modelname{} also achieves consistently lower perplexity than the MoE baseline at every scale, with $-12.2\%$ at S, $-11.7\%$ at M, and $-18.8\%$ at L, suggesting that the quality of language modeling improves consistently with scale under \modelname{}'s design, independent of the downstream accuracy results.

Taken together, the statistical analysis supports the conclusion that \modelname{} produces a reliable improvement over the standard MoE baseline. The effect is consistent in direction across all three scales and nine benchmarks, reaches statistical significance when observations are pooled with paired $t$-test $p = 0.037$, and is accompanied by a moderate effect size with Cohen's $d = 0.36$, along with substantial perplexity reduction at every scale.

\section{Additional Discussion}

\subsection{Load-Balancing}

A recent line of work explored auxiliary-loss-free load balancing \citep{liu2024deepseek, guo2025deepseek}.
Our baselines and Routing-Free MoE all employ auxiliary losses following standard practice, ensuring a controlled comparison across routing architectures.
Since auxiliary-loss-free balancing is orthogonal to the routing mechanism itself, any such technique could in principle be applied to standard MoE, AoE, ReMoE, and Routing-Free MoE alike.
Introducing it for one method but not others would confound the comparison, while applying it uniformly across all baselines would constitute a substantial engineering effort that lies beyond the scope of this work.
We therefore focus on the architectural distinction between routing-based and routing-free expert selection, which is the central contribution of this paper, and leave the integration of auxiliary-loss-free balancing as beyond the scope of this work.

Similarly, since all architectures compared in this work operate under the Token Choice load balancing paradigm, which is the dominant setting in the MoE literature \citep{muennighoff2024olmoe}, we choose not to include Expert Choice routing as a direct baseline.
Instead, our configurable $\mu$-interpolation between token-balancing and expert-balancing losses allows Routing-Free MoE to smoothly interpolate the spectrum between token-balancing and expert-balancing behavior within a unified framework.
This design choice enables a thorough internal comparison across balancing strategies, as demonstrated in our ablation studies, without introducing the confound of an entirely different assignment paradigm.

\subsection{Per-Layer and Global Density}
\label{appendix:density}

In Figure \ref{fig:density_layer}, without any explicit supervision, the model develops a striking three-stage structure, with early layers showing high but rapidly decreasing activation with large variance, middle layers exhibiting a slow monotonic climb with low variance, and late layers displaying a sharp rise in both activation level and variance.
This emergent pattern coincides closely with interpretability findings on the heterogeneous functional roles that layers naturally develop in LLMs \citep{geva2021transformer, gao2024mola, wendler2024llamas, jiao2024spin}, with early layers converting input from token space to concept space, middle layers processing them, and late layers preparing input-dependent outputs back to token space.
We argue that the performance gain from global density constraints stems precisely from removing the per‑layer inductive bias. 
Enforcing identical sparsity at every depth suppresses the compute‑hungry layers that naturally benefit from activating more experts, while simultaneously forcing unnecessary activations in layers where sparse representations suffice. Once this bias is lifted, the model is free to self‑organize into a more effective, functionally aligned activation structure.

\rmv{
\begin{figure}[t]
    \centering
    \includegraphics[width=0.6\linewidth]{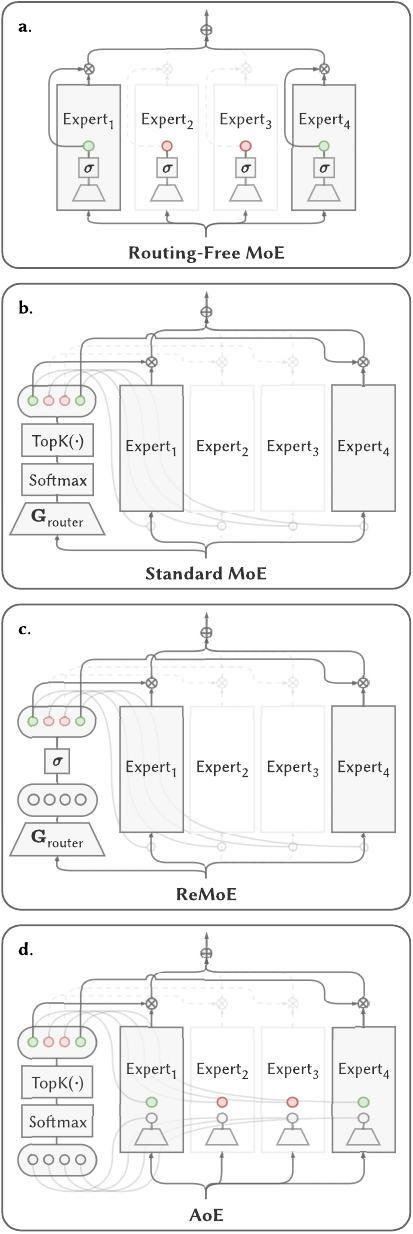}
    \caption{Architectural comparison between (a) Routing-Free MoE; (b) standard MoE \citep{fedus2022switchtransformer}; (c) ReMoE \citep{wang2024remoe}; (d) AoE \citep{lv2025aoe}. Routing-Free MoE encapsulates all sparse-activation mechanisms within individual experts, enabling each expert to determine its activation entirely on its own. }
    \label{fig:routing-free-moe}
\end{figure}
}

\section{Additional Experiment Results}\label{appdix:result}

\begin{table*}[htbp]
\caption{\textbf{Performance of standard MoE, AoE, ReMoE, and \modelname{} across configurations.} Each model was trained on OpenWebText \citep{Gokaslan2019OpenWeb}. 
Entries marked as n/a indicate fields not applicable to the respective architectural design. 
Results underlined denote experiments with gradient explosion.}
\label{tab:all-result}
\vspace{0.8em}

\begin{adjustbox}{width=\textwidth}
    \centering
    \addtolength{\tabcolsep}{-2.5pt} 
    \small
    \begin{tabular}{rc|ccccrr|clcc|ccc}
    \toprule
        \textbf{Arch.} &
        \textbf{Config.} &
        \textbf{$r$} &
        \textbf{$L$} &
        \textbf{$D$} &
        \textbf{$D_\mathrm{act}$} &
        \textbf{Size} & \textbf{FLOPs} &
        \textbf{$\lambda$} &
        \textbf{$\eta$} &
        \textbf{$\mu$} &
        \textbf{$\alpha$} &
        \textbf{Loss} &
        \textbf{Val. Loss} &
        \textbf{PPL} \\
    \midrule
        MoE   & Top3/12 & n/a & 12 & 512 & 128 & 92.44M & 90.93M & n/a & n/a & n/a & 5e${}^{-4}$ & 3.804 & 3.667 & 39.13 \\
        MoE   & Top3/12 & n/a & 12 & 512 & 128 & 92.44M & 90.93M & n/a & n/a & n/a & 1e${}^{-3}$ & {3.595} & {3.441} & {31.22}\\
        MoE   & Top3/12 & n/a & 12 & 512 & 128 & 92.44M & 90.93M & n/a & n/a & n/a & 2e${}^{-3}$ & \underline{5.584} & \underline{5.366} & \underline{214.0} \\
    \midrule
        AoE   & Top3/12 & 16 & 12 & 512 & 128 & 93.85M & 88.57M & n/a & n/a & n/a & 1e${}^{-3}$ & 3.440 & 3.401 & 30.00 \\
        AoE   & Top3/12 & 32 & 12 & 512 & 128 & 95.32M & 91.08M & n/a & n/a & n/a & 1e${}^{-3}$ & 3.450 & 3.411 & 30.31 \\
        ReMoE & Top3/12 & n/a & 12 & 512 & 128 & 92.44M & 90.93M & 1e${}^{-8}$ & 0.2 & n/a & 1e${}^{-3}$ & 3.521 & 3.389 & 29.60 \\
    \midrule
        RFMoE   & $\rho_\infty$=1/4 & 16 & 12 & 512 & 128 & 93.85M & 88.57M & 1e${}^{-10}$ & 0.005 & 0.5 & 5e${}^{-4}$ & 3.579 & 3.620 & 37.34 \\  
        RFMoE   & $\rho_\infty$=1/4 & 16 & 12 & 512 & 128 & 93.85M & 88.57M & 1e${}^{-10}$ & 0.05 & 0.5 & 5e${}^{-4}$ & 3.552 & 3.600 & 36.60 \\
        RFMoE   & $\rho_\infty$=1/4 & 16 & 12 & 512 & 128 & 93.85M & 88.57M & 1e${}^{-10}$ & 0.08 & 0.5 & 5e${}^{-4}$ & 3.537 & 3.644 & 38.24 \\
        RFMoE   & $\rho_\infty$=1/4 & 16 & 12 & 512 & 128 & 93.85M & 88.57M & 1e${}^{-10}$ & 0.1 & 0.5 & 5e${}^{-4}$ & 3.571 & 3.612 & 37.04 \\

    \midrule
        RFMoE   & $\rho_\infty$=1/4 & 16 & 12 & 512 & 128 & 93.85M & 88.57M & 1e${}^{-10}$ & 0.02 & 0.5 & 1e${}^{-3}$ & 3.309 & 3.358 & 28.73 \\
        RFMoE   & $\rho_\infty$=1/4 & 16 & 12 & 512 & 128 & 93.85M & 88.57M & 1e${}^{-10}$ & 0.02 & 0.5 & 2e${}^{-3}$ & {3.261} & {3.311} & {27.42} \\
        RFMoE   & $\rho_\infty$=1/4 & 16 & 12 & 512 & 128 & 93.85M & 88.57M & 1e${}^{-10}$ & 0.02 & 0.5 & 5e${}^{-3}$ & \underline{7.097} & \underline{6.971} & \underline{1065}\\

    \midrule
        RFMoE   & $\rho_\infty$=1/4 & 8 & 12 & 512 & 128 & 93.11M & 87.32M & 1e${}^{-10}$ & 0.02 & 0.5 & 1e${}^{-3}$ & 3.332 & 3.373 & 29.16 \\
        RFMoE   & $\rho_\infty$=1/4 & 16 & 12 & 512 & 128 & 93.85M & 88.57M & 1e${}^{-10}$ & 0.02 & 0.5 & 1e${}^{-3}$ & 3.309 & 3.358 & 28.74 \\
        RFMoE   & $\rho_\infty$=1/4 & 32 & 12 & 512 & 128 & 95.32M & 91.08M & 1e${}^{-10}$ & 0.02 & 0.5 & 1e${}^{-3}$ & 3.290 & 3.344 & 28.34 \\
        RFMoE   & $\rho_\infty$=1/4 & 64 & 12 & 512 & 128 & 98.27M & 96.09M & 1e${}^{-10}$ & 0.02 & 0.5 & 1e${}^{-3}$ & 3.290 & 3.341 & 28.24 \\

    \midrule
        RFMoE   & $\rho_\infty$=1/4 & 8 & 12 & 512 & 128 & 93.11M & 87.32M & 1e${}^{-10}$ & 0.02 & 0.5 & 2e${}^{-3}$ & 3.717 & 3.699 & 40.41 \\
        RFMoE   & $\rho_\infty$=1/4 & 16 & 12 & 512 & 128 & 93.85M & 88.57M & 1e${}^{-10}$ & 0.02 & 0.5 & 2e${}^{-3}$ & 3.311 & 3.261 & 26.08 \\
        RFMoE   & $\rho_\infty$=1/4 & 32 & 12 & 512 & 128 & 95.32M & 91.08M & 1e${}^{-10}$ & 0.02 & 0.5 & 2e${}^{-3}$ & \underline{6.435} & \underline{6.564} & \underline{709.1} \\
        RFMoE   & $\rho_\infty$=1/4 & 64 & 12 & 512 & 128 & 98.27M & 96.09M & 1e${}^{-10}$ & 0.02 & 0.5 & 2e${}^{-3}$ & 3.548 & 3.516 & 33.65 \\

    \midrule
        RFMoE   & $\rho_\infty$=1/4 & 32 & 12 & 512 & 128 & 95.32M & 91.08M & 1e${}^{-10}$ & 0.02 & 0.0 & 1e${}^{-3}$ & 3.481 & 3.347 & 28.41 \\
        RFMoE   & $\rho_\infty$=1/4 & 32 & 12 & 512 & 128 & 95.32M & 91.08M & 1e${}^{-10}$ & 0.02 & 0.2 & 1e${}^{-3}$ & 3.483 & 3.345 & 28.35 \\
        RFMoE   & $\rho_\infty$=1/4 & 32 & 12 & 512 & 128 & 95.32M & 91.08M & 1e${}^{-10}$ & 0.02 & 0.8 & 1e${}^{-3}$ & 3.488 & 3.346 & 28.38 \\
        RFMoE   & $\rho_\infty$=1/4 & 32 & 12 & 512 & 128 & 95.32M & 91.08M & 1e${}^{-10}$ & 0.02 & 1.0 & 1e${}^{-3}$ & 3.485 & 3.347 & 28.43 \\

    \midrule
    \midrule
        MoE   & Top4/16 & n/a & 24 & 768 & 192 & 289.92M & 247.95M & n/a & n/a & n/a & 5e${}^{-4}$ & 3.808 & 3.300 & 27.11 \\
        MoE   & Top4/16 & n/a & 24 & 768 & 192 & 289.92M & 247.95M & n/a & n/a & n/a & 1e${}^{-3}$ & {3.726} & {3.219} & {25.00} \\
    \midrule
        RFMoE   & $\rho_\infty$=1/4 & 48 & 24 &  768 & 192 & 307.30M & 249.17M & 1e${}^{-10}$ & 0.02 & 0.5 & 5e${}^{-4}$ & 3.360 & 3.292 & 26.90 \\
        RFMoE   & $\rho_\infty$=1/4 & 48 & 24 &  768 & 192 & 307.30M & 249.17M & 1e${}^{-10}$ & 0.02 & 0.5 & 1e${}^{-3}$ & {3.166} & {3.095} & {22.08} \\
    \midrule
    \midrule
        MoE   & Top6/24 & n/a & 32 & 1024 & 256 & 808.42M & 608.38M & n/a & n/a & n/a & 5e${}^{-4}$ & {3.947} & {3.202} & {24.58} \\
        MoE   & Top6/24 & n/a & 32 & 1024 & 256 & 808.42M & 608.38M & n/a & n/a & n/a & 8e${}^{-4}$ & \underline{4.493} & \underline{3.768} & \underline{43.29} \\
        MoE   & Top6/24 & n/a & 32 & 1024 & 256 & 808.42M & 608.38M & n/a & n/a & n/a & 1e${}^{-3}$ & \underline{8.532} & \underline{8.104} & \underline{3309} \\
    \midrule
        RFMoE   & $\rho_\infty$=1/4 & 64 & 32 & 1024 & 256 & 870.60M & 613.19M & 1e${}^{-10}$ & 0.02 & 0.5 & 5e${}^{-4}$ & 3.179 & 3.107 & 22.36 \\
        RFMoE   & $\rho_\infty$=1/4 & 64 & 32 & 1024 & 256 & 870.60M & 613.19M & 1e${}^{-10}$ & 0.02 & 0.5 & 8e${}^{-4}$ & {3.076} & {2.994} & {19.97} \\
        RFMoE   & $\rho_\infty$=1/4 & 64 & 32 & 1024 & 256 & 870.60M & 613.19M & 1e${}^{-10}$ & 0.02 & 0.5 & 1e${}^{-3}$ & \underline{5.448} & \underline{3.548} & \underline{34.73} \\
    \bottomrule
    \end{tabular}
\end{adjustbox}
\end{table*}

\end{document}